\documentclass[]{fairmeta}

\usepackage{graphicx}
\usepackage{wrapfig}
\usepackage{tabularx}
\usepackage{booktabs}
\usepackage{multirow}
\usepackage{makecell}
\usepackage{colortbl}
\usepackage[dvipsnames]{xcolor}
\usepackage{textcomp}
\usepackage{url}
\usepackage{verbatim}
\usepackage{titlesec}
\usepackage{tocloft}
\usepackage{adjustbox}
\usepackage{subcaption}
\usepackage{float}
\usepackage{stfloats}
\usepackage{siunitx}
\usepackage{microtype}
\usepackage{tikz}
\usepackage{svg}
\usepackage{animate}
\usepackage{etoc}
\usepackage{comment}

\usepackage{amsmath,amssymb}

\usepackage{enumitem}
\usepackage{tcolorbox}
\usepackage{ragged2e}
\tcbset{
  promptbox/.style={
    colback=black!2,
    colframe=black!20,
    boxrule=0.3pt,
    arc=1mm,
    left=6pt,
    right=6pt,
    top=6pt,
    bottom=6pt
  }
}

\usepackage{algorithm}
\usepackage{algpseudocode}

\algrenewcommand{\algorithmiccomment}[1]{\hfill\textcolor{light_blue}{// #1}}

\usepackage{pifont}
\usepackage{bbding}
\usepackage{xspace}

\definecolor{adptorange}{RGB}{248,205,172}
\definecolor{cmpblue}{RGB}{189,215,238}

\definecolor{our_red}{RGB}{232,157,160}
\definecolor{our_blue}{RGB}{136,206,230}
\definecolor{our_orange}{RGB}{246,200,168}
\definecolor{our_green}{RGB}{178,211,164}

\definecolor{attn_code0}{RGB}{247,215,200}
\definecolor{attn_code1}{RGB}{238,169,139}
\definecolor{mlp_code0}{RGB}{204,201,221}
\definecolor{mlp_code1}{RGB}{102,95,153}

\definecolor{dark_green}{rgb}{0,0.5,0}
\definecolor{dark_red}{rgb}{0.8,0.2,0.2}
\definecolor{soft_red}{rgb}{1.0,0.4,0.4}
\definecolor{light_blue}{rgb}{0.2,0.5,1.0}

\definecolor{token_blue}{RGB}{84,120,140}
\definecolor{darkgreen}{rgb}{0.15,0.75,0.15}
\definecolor{cvprblue}{rgb}{0.21,0.49,0.74}
\definecolor{lightblue}{rgb}{0.90,0.95,0.99}

\usepackage{hyperref}
\usepackage{cleveref}

\usepackage{natbib}

\makeatletter
\DeclareRobustCommand\onedot{\futurelet\@let@token\@onedot}
\def\@onedot{\ifx\@let@token.\else.\null\fi\xspace}

\makeatother

\title{
  BIFE: Better Interaction, Fewer Errors for Minute-Long Video Generation
}

\author[1*]{Zeyu Zhang}
\author[2*]{Jinyuan Mao}
\author[12]{Shuning Chang}
\author[12]{Yuanyu He}
\author[1]{Yizeng Han}
\author[13\dag]{Jiasheng Tang}
\author[1]{Fan Wang}
\author[12\dag]{Bohan Zhuang}

\affiliation[1]{DAMO Academy, Alibaba Group}
\affiliation[2]{Zhejiang University}
\affiliation[3]{Hupan Lab}
\contribution[*]{Equal contribution.}
\contribution[\dag]{Corresponding authors.}

\abstract{
    Long video generation is a critical step toward building realistic world models, requiring both high visual fidelity and long-range interaction consistency. Recent autoregressive diffusion models enable long-horizon generation via KV cache reuse, yet suffer from two fundamental challenges: \emph{fail to preserve long-range interactions} caused by sliding-window KV cache and \emph{error accumulation} that progressively degrades generation quality over time. To address these issues, we propose \textbf{BIFE}, a framework that introduces a \emph{semantic sparse KV cache} for retrieval-based long-range conditioning and a \emph{Block Forcing} training strategy to enforce cross-block consistency. Together, these designs preserve historical interactions while mitigating drift, enabling stable and coherent minute-long video generation. We also introduce \textbf{InterVBench}, a minute-long video benchmark with fine-grained block-level annotations and Video Drift Error metrics. Extensive experiments on InterVBench and VBench-Long demonstrate that BIFE achieves state-of-the-art performance, including a \textbf{22.2\%} improvement on VDE-Subject and a \textbf{19.4\%} improvement on VDE-Clarity over baselines.}

\date{\today} 
\metadata[Website]{\url{https://alibaba-damo-academy.github.io/BIFE/}}
\metadata[Code]{\url{https://github.com/alibaba-damo-academy/BIFE/}}
\metadata[Correspondence]{\email{jiasheng.tjs@alibaba-inc.com}, \email{bohan.zhuang@gmail.com}}

\begin{document}
\thispagestyle{firstheader}
\maketitle

\section{Introduction}

\begin{figure*}[b]
    \centering
    \includegraphics[width=\linewidth]{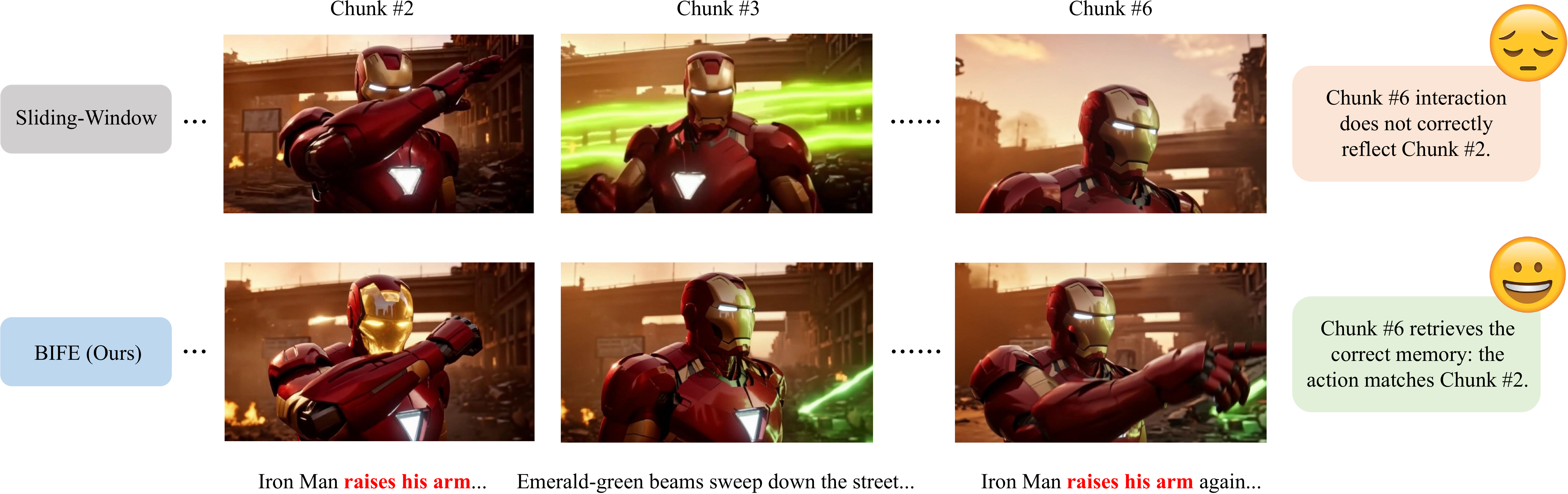}
    \caption{\textbf{Sliding-Window vs. Sparse Retrieval in KV cache.} Existing methods (top) fail to preserve long-range interactions, causing later chunks (e.g., chunk 6) to drift from earlier states (chunk 2), while ours (bottom) retrieves relevant KV cache and maintain long-term memory during interactive generation.}
    \label{fig:teaser}
\end{figure*}

Long video generation is crucial for creating realistic and coherent narratives that unfold over extended durations, essential for filmmaking, digital storytelling, and virtual simulation~\cite{yi2025magic,wang2025lingen,huang2024owl,liu2025fpsattention}. Moreover, generating minute-long videos is a key step toward world models, which act as foundational simulators for agentic AI, embodied AI, and gaming~\cite{chegamegen,shi2025presentagent}.
Minute-long video generation requires balancing fidelity, coherence, and efficiency.
The autoregressive diffusion paradigm~\cite{arriola2025block,huang2025self,teng2025magi} provides a promising foundation to address these issues by generating video tokens in blocks, performing diffusion-based denoising within each block while conditioning causally on previously generated ones. By introducing KV cache management such as sliding-window KV cache~\cite{huang2025self,teng2025magi} into diffusion models, autoregressive diffusion enables efficient, variable-length, and parallelizable generation without sacrificing per-block visual fidelity.

\begin{figure*}[t]
    \centering
    \includegraphics[width=\linewidth]{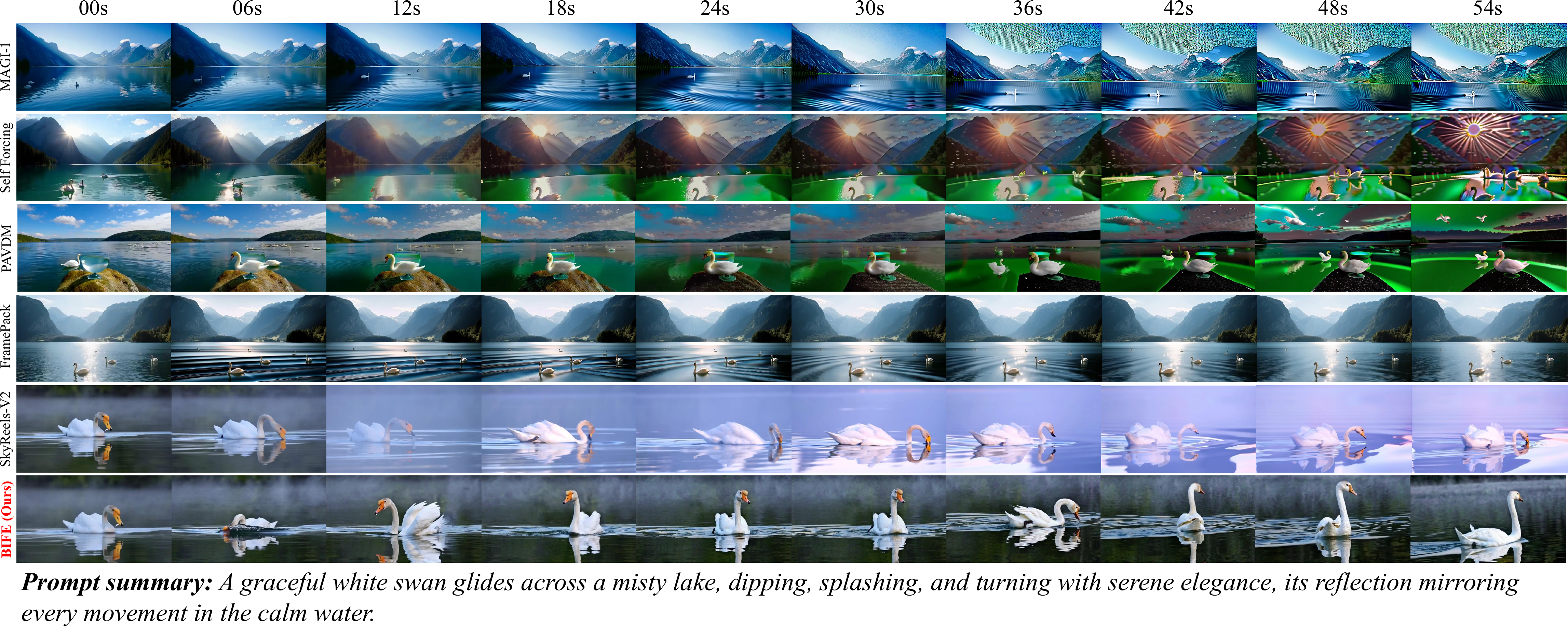}
    \caption{\textbf{Visualized comparison between \textit{BIFE} and baselines in terms of accumulation error.} See more details in \textit{Appendix: Visualization Comparison}.}
    \label{fig:demo}
\end{figure*}

However, existing autoregressive diffusion models still face three fundamental challenges. First, the existing \textbf{\emph{sliding-window KV cache loses long-range historical information and cannot preserve previous interactions}}. Once the cache reaches the window size, earlier tokens are dropped to accommodate new ones, thereby permanently losing historical information. As the concrete example shown in Fig.~\ref{fig:teaser}, previous works~\cite{liu2025rolling,cui2025self} fail to maintain long-range memory, leading to significant deviation in later chunks (e.g., chunk 6) from earlier interactions (e.g., chunk 2).
Second, \textbf{\emph{the AR paradigm inevitably suffers from error accumulation}}, where small prediction errors gradually build up over time and will be directly stored in the KV cache~\cite{kang2024gear}. In long video generation, the accumulated errors typically manifest as quality degradation, color drift, subject and background inconsistency, and visual distortions~\cite{lu2024freelong}, as shown in Fig.~\ref{fig:demo}. As a result, existing sliding-window KV cache~\cite{huang2025self,teng2025magi} is insufficient for stable, minute-long generation, necessitating principled mechanisms to explicitly mitigate error accumulation.
Third, the domain is hindered by \textbf{\emph{the lack of long video benchmarks with block-level interactive prompts}}. Currently, most open-source long-video benchmarks are either video-level annotations~\cite{nan2024openvid} or non-continuous fragments~\cite{wu2025moviebench}. Meanwhile, existing standard benchmarks such as VBench~\cite{huang2024vbench,huang2025vbench++,zheng2025vbench} focus on diversity or object categories but fail to capture error accumulation and coherence over extended durations.

To address these challenges, we propose \textbf{\textit{BIFE}}, an autoregressive diffusion model designed for \textbf{B}etter \textbf{I}nteraction and \textbf{F}ewer \textbf{E}rrors for minute-long video generation through innovative design in both memory and training strategies.
In terms of memory, we propose a tailored KV cache management strategy termed \textbf{\textit{semantic sparse KV cache}}. It stores the sparse KV cache for all previous blocks and retrieves the most semantically aligned context for the current prompt, thereby accurately reproducing previous interactions and maintaining long-range consistency without propagating redundant errors.
In terms of training, we propose \textbf{\textit{Block Forcing}}, that mitigates the long-range training-inference gap by explicitly regularizing cross-block coherence.
This prevents models from drifting over long horizons, such as losing track of subjects or gradually altering scene content. 

To address the lack of interactive long-video benchmarks, we propose \emph{InterVBench}, comprising 1,000 minute-long videos annotated every 2–5 seconds. To better evaluate long video generation quality, we further introduce Video Drift Error (VDE) metrics based on Weighted Mean Absolute Percentage Error (WMAPE)~\cite{kim2016new,de2016mean}, integrated with original VBench metrics, providing a more comprehensive reflection of long-horizon interaction, temporal consistency, and visual fidelity.

Comprehensive experiments are conducted on both InterVBench and the traditional VBench to demonstrate the superiority of our method.
Notably, \textit{BIFE} outperforms the state of the art on InterVBench, with improvements of \textbf{22.2\%} on VDE-Subject and \textbf{19.4\%} on VDE-Clarity.

Our contributions can be summarized as follows:

\begin{itemize}
    \item We present \textbf{BIFE}, an autoregressive diffusion framework for interactive minute-long video generation. It introduces a \textit{semantic-aware sparse KV cache}
    along with a novel \textit{Block Forcing} training strategy. These components jointly improve interaction and reduce error accumulation in minute-long video generation.

    \item We introduce \textbf{InterVBench}, a benchmark of 1,000 minute-long videos with %
    block-level annotations, along with the \textit{Video Drift Error} (VDE) metric to evaluate long-horizon interaction, temporal consistency, and visual fidelity.
    
    \item We conduct extensive experiments on InterVBench and VBench, demonstrating that BIFE significantly outperforms state-of-the-art baselines across both interaction and visual quality metrics.
\end{itemize}

\section{Related Work}

\begin{figure*}[t]
\centering

\begin{minipage}[t]{0.62\textwidth}
    \vspace{0pt}
    \centering
    \includegraphics[width=\linewidth]{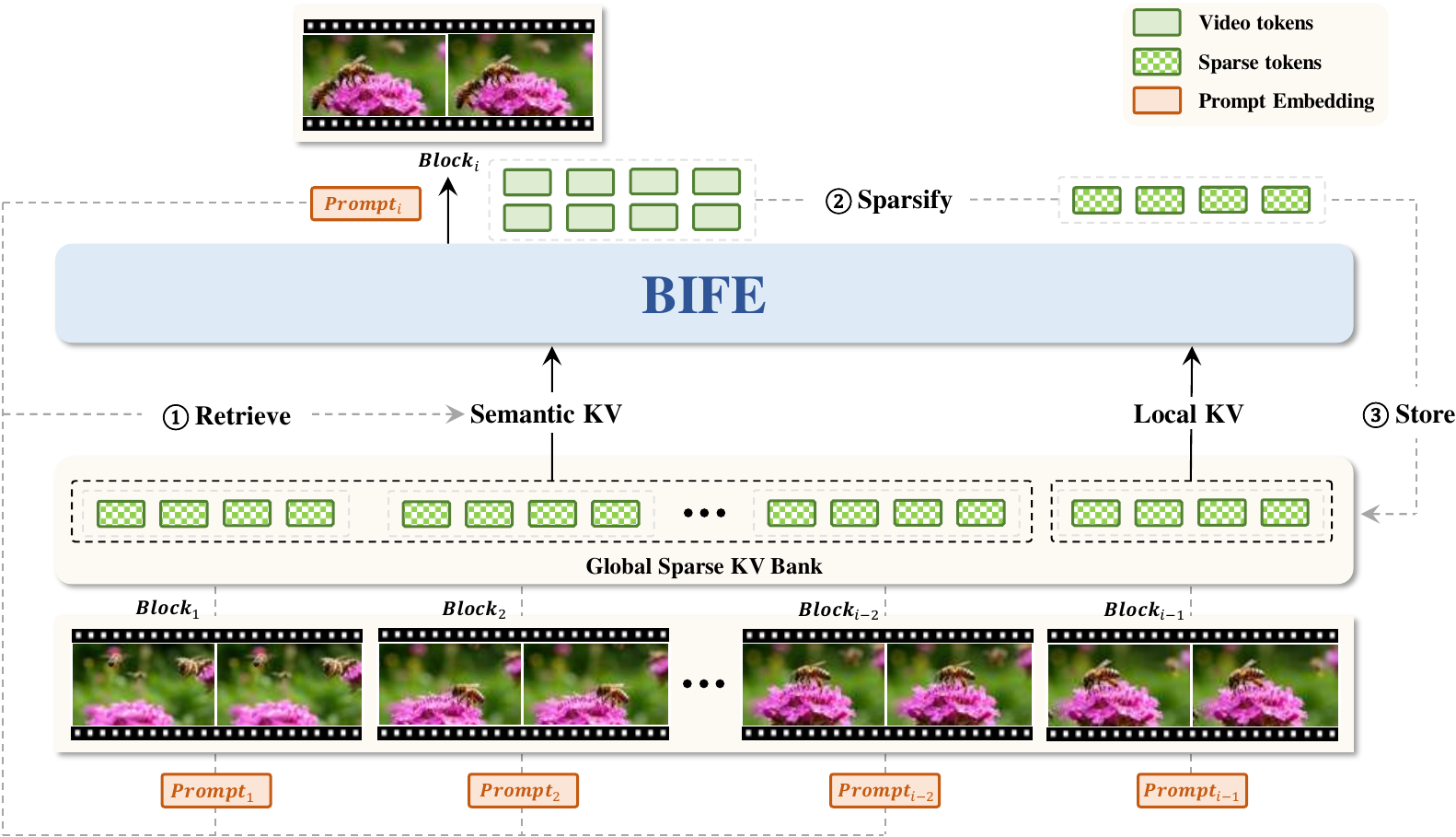}
    \caption{\textbf{Overview of the BIFE framework.} The generation of block $i$ is conditioned on both a local KV cache and a globally retrieved context. The global context is dynamically assembled by retrieving top-$l$ semantically similar KV blocks via prompt embedding similarity. Upon generation, the global sparse KV bank is updated with the new block's most salient KV tokens.}
    \label{fig:arch}
\end{minipage}
\hfill
\begin{minipage}[t]{0.365\textwidth}
    \vspace{0pt}
    \centering
    \includegraphics[width=\linewidth]{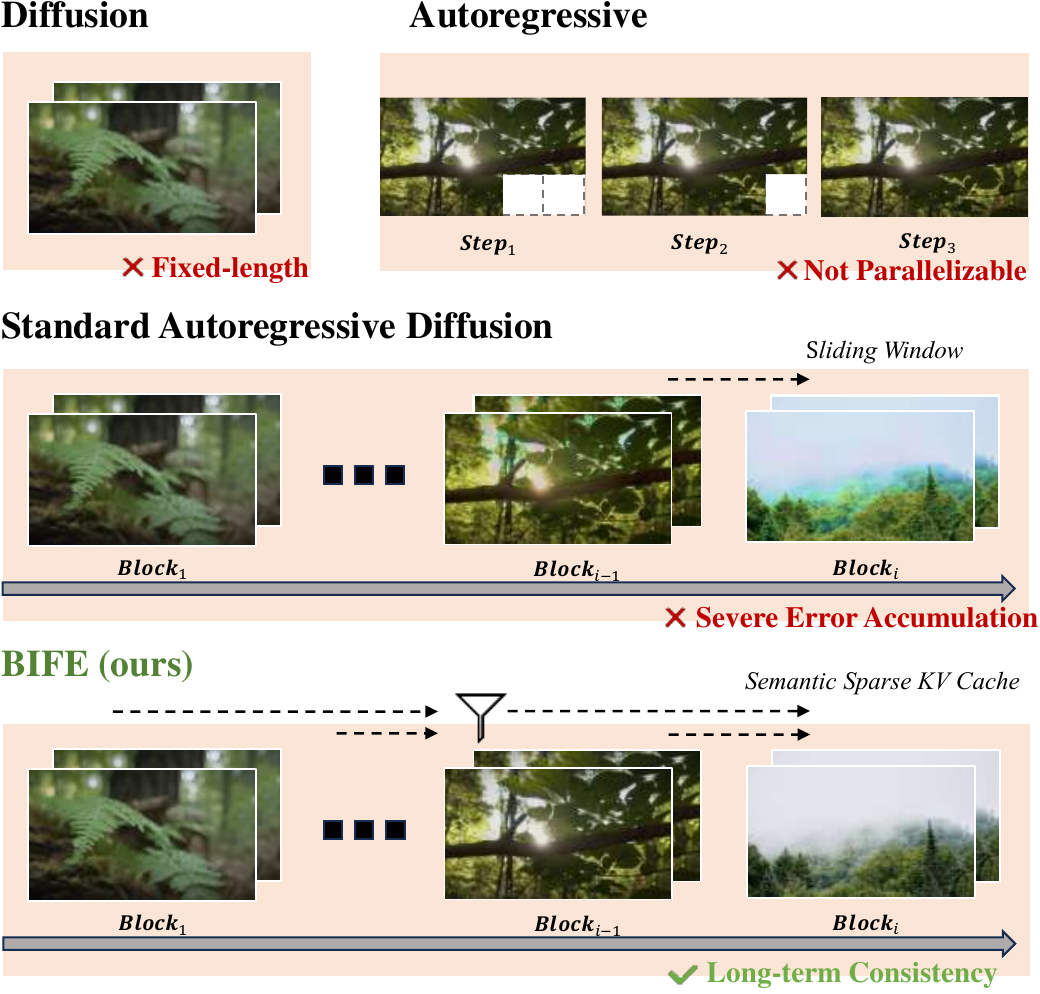}
    \vskip -0.1in
    \caption{\textbf{Architecture comparison: AR vs. Diffusion vs. Autoregressive Diffusion vs. BIFE (ours)}. Our \textbf{BIFE} addresses long-horizon interaction and block-wise error accumulation, enabling high-fidelity and coherent minute-long video generation.}
    \label{fig:arch-compare}
\end{minipage}
\vspace{-1em}
\end{figure*}

\textit{Minute-long video generation} can be roughly grouped into three settings: single-shot, multi-shot generation, and movie-style video composition.
(1) Single-shot generation aims to produce a minute-long block within a consistent scene and semantic context, emphasizing long-range temporal coherence and visual stability. Approaches fall into AR and autoregressive diffusion families.
AR methods, such as FAR~\cite{gu2025long} and Loong~\cite{wang2024loong}, formulate long video generation as next-frame (or next segment) prediction. 
autoregressive diffusion models generate videos block by block, while performing iterative diffusion-based~\cite{peebles2023scalable} denoising within each block.
Their key design choice lies in the block-level causal conditioning: MAGI-1~\cite{teng2025magi}, Skyreel-V2~\cite{chen2025skyreels}, Self Forcing~\cite{huang2025self}, Rolling Forcing~\cite{liu2025rolling}, and LongLive~\cite{yang2025longlive} proceed strictly sequentially across blocks, whereas FramePack~\cite{zhang2025packing} adopts a symmetric schedule that treats both ends as guidance and fills the middle autoregressively. In practice, autoregressive diffusion models methods typically rely on \emph{KV cache design} for efficiency and stability over long horizons.
As shown in Fig.~\ref{fig:arch-compare}, diffusion-based video models, typically instantiated with Diffusion Transformers (DiT)~\cite{peebles2023scalable,wan2025wan}, employ bidirectional attention without KV caching, which enables strong local consistency and controllability but leads to inefficient decoding and rigid fixed-length generation. Autoregressive (AR) frameworks~\cite{wang2024loong}, on the other hand, naturally support variable-length decoding with KV cache reuse, yet suffer from limited parallelism and degraded visual quality when extended to long temporal horizons. These limitations become particularly severe in minute-scale generation, where even small prediction errors can accumulate and amplify over time. 
(2) Multi-shot generation typically focuses on handling camera motions and transitions across scenes or semantics. Recent systems, such as LCT~\cite{guo2025long}, RIFLEx~\cite{zhao2025riflex}, and MoC~\cite{cai2025mixture}, organize text–video units with interleaved layouts and positional extrapolation to accommodate multiple shots.
(3) Movie-style generation aims to create cinematic content by stitching multiple blocks with different scenes and styles, while maintaining a coherent global narrative or theme. Methods~\cite{dalal2025one,zhao2024moviedreamer,wu2025moviebench,xiao2025captain} resemble film editing, combining diverse shots into a single coherent video guided by block-level text descriptions.

\textit{Autoregressive diffusion model} decodes long sequences in blocks: within each block the model performs iterative diffusion denoising, while across blocks it conditions on previously generated content via KV caches. This paradigm has been explored in both text and video.
In language modeling, BD3-LM~\cite{arriola2025block} and SSD-LM~\cite{han2022ssd} demonstrate that block-wise diffusion can combine bidirectional refinement within a block with efficient, variable-length decoding through cached context across blocks. In video generation, related formulations adopt autoregressive diffusion with causal conditioning to interpolate between pure diffusion (e.g., DiT-style bidirectional attention without KV caching) and AR (variable-length decoding with KV caching but weaker visual fidelity and limited parallelism). Representative models include MAGI-1~\cite{teng2025magi}, Self Forcing~\cite{huang2025self}, Self-Forcing++~\cite{cui2025self}, Rolling Forcing~\cite{liu2025rolling}, Deep Forcing~\cite{yi2025deep}, StreamDiffusionV2~\cite{feng2025streamdiffusionv2}, PAVDM~\cite{xie2025progressive}, LongLive~\cite{yang2025longlive}, CausVid~\cite{yin2025causvid}, SkyReels-V2~\cite{chen2025skyreels}, and Infinity-RoPE~\cite{yesiltepe2025infinity}, etc, which condition each new block on past blocks to extend temporal horizons while retaining diffusion’s denoising quality within a block.
Despite progress, autoregressive diffusion models remain constrained by long-horizon information loss, KV cache–induced errors, and the lack of interactive long video benchmarks. We address these gaps with (1) \textit{BIFE}, a framework featuring semantic sparse KV cache and Block Forcing to enhance long-range interactive generation, and (2) \textit{InterVBench}, a benchmark of 1,000 minute-long videos with block-level interactive prompts for evaluation.

\section{Preliminaries}

\subsection{Flow Matching} 

From a fidelity perspective, \textit{BIFE} preserves the reconstruction quality of the current block by supervising the model under the stochastic interpolant formulation~\cite{albergo2022building} of Flow Matching~\cite{lipman2022flow}. Instead of predicting additive Gaussian noise as in DDPM, Flow Matching constructs a continuous trajectory between the real starting frame \(x_{\text{start}}\) and a Gaussian endpoint \(\epsilon \sim \mathcal{N}(0,I)\) at time $t$:
\begin{equation}
x_t = (1 - t)x_{\text{start}} + t\epsilon.
\end{equation}
Along this trajectory, the model predicts the corresponding velocity field
\begin{equation}
v_t = \epsilon - x_{\text{start}},
\end{equation}
which provides the denoising direction in latent space and ensures accurate reconstruction of the current block.

In the stochastic interpolant formulation of flow matching~\cite{albergo2022building}, the model predicts a velocity field $v_{\text{pred}}$ that represents the temporal derivative of the interpolated state $x_t$ between noise and data: 
\begin{equation}
v_{\text{pred}}= v_t(x_t) = \frac{d}{dt} x_t = f_{\theta}(x_t, t).
\end{equation}

\subsection{Self Forcing}

Long video generation suffers from the \textit{training-inference gap}: during training the model conditions on ground-truth frames (teacher forcing), but at inference it relies on its own imperfect outputs, leading to exposure bias and error accumulation. To address this, we employ  \textit{Self Forcing}~\cite{huang2025self}, where the model generates a full video sequence $\tilde{x}_{1:T}$ semi-autoregressively and is then penalized at the \emph{video level} by matching its distribution $p_\theta$ to the real distribution $p_{\text{data}}$. Concretely, a discriminator $D$ evaluates entire videos, and the generator $G$ is trained to minimize 
\begin{equation}
\begin{aligned}
    \mathcal{L}_{\text{SF}} &= \min_{G}\max_{D}\;\;  \mathbb{E}_{x\sim p_{\text{data}}}\![\log D(x)] \\
    &+ \mathbb{E}_{\tilde{x}\sim p_\theta}\![\log(1-D(\tilde{x}))],
\end{aligned}
\end{equation}
where $\tilde{x}\sim p_\theta$ is obtained by the predictions of $G$. 
This formulation exposes the model to its own errors during training and enforces sequence-level realism, thereby reducing exposure bias and improving temporal consistency.

\section{Method}

\subsection{Overview}

Long-horizon autoregressive diffusion suffers from both long-range information loss and error accumulation during interactive generation. These problems arise from two sources: one from memory, where earlier tokens are dropped when new tokens are appended to the KV cache, and block-level errors are propagated through KV cache conditioning; and the other from training, where exposure bias leads to cross-block drift. \textit{BIFE} explicitly addresses both by combining KV cache memory design for interactive long-range conditioning (Section~\ref{sec:KV}) and training objectives for cross-block regularization (Section~\ref{sec:bf}).

\textit{BIFE} introduces an autoregressive diffusion architecture, as illustrated in Fig.~\ref{fig:arch}. At inference, the model generates an interactive long video
$V = \{V_1, V_2, V_3, \ldots, V_n\}$ in a block-by-block manner, where each block
$V_i \in \mathbb{R}^{(1+T) \times H \times W \times 3}$ consists of $T$ frames with spatial resolution $H \times W$ and three RGB channels. The generation of each block is conditioned on the corresponding block-level prompt
$\mathcal{Y}=\{y_i\}_{i=1}^{n}$, where $y_i$ provides semantic guidance for block $V_i$. 

To enable interactive long-range conditioning during inference, \textit{BIFE} maintains a semantic sparse KV cache across blocks. Specifically, the key–value pairs produced during the generation of previous blocks are selectively stored in a global KV bank. For each new block, a small set of semantically relevant KV entries is retrieved from the bank based on prompt similarity and combined with recent local context, forming the conditional context for the current block generation. This sparse retrieval mechanism limits error propagation while preserving semantically aligned long-range dependencies.

Autoregressive diffusion is performed in the latent space. Each video block is represented by a latent tensor
$Z_i \in \mathbb{R}^{(1 + T/4) \times H/8 \times W/8 \times 16}$, which is first processed by the autoregressive diffusion denoiser to produce a denoised latent $\tilde{Z_i}$. The denoising process operates under causal conditioning across blocks and leverages the retrieved KV context to guide long-horizon generation. After denoising, $\tilde{Z_i}$ is decoded by a 3D causal VAE decoder to reconstruct video frames $V_i$ in the original video space. 
The decoder restores both spatial and temporal resolution, with the first frame decoded using spatial upsampling only to preserve image-guidance fidelity.

During post-training, we further introduce Block Forcing, a training objective that alleviates the training–inference discrepancy by enforcing inter-block consistency and conditioning on the model’s own latent trajectories, extending the principles of Self Forcing.

\subsection{Semantic Sparse KV Cache}
\label{sec:KV}

Long video generation requires preserving dependencies across many blocks. However, storing and conditioning on the full KV context imposes heavy memory, computational burdens, and accumulation error. Moreover, simply caching the most recent blocks fails to capture long-range semantic relations. To address this, we introduce \textit{Semantic Sparse KV Cache} that selectively stores only the most informative tokens and retrieves relevant past
KV blocks, enabling efficient and interactive long-range conditioning.

We first dynamically identify salient tokens with a probing mechanism and store the most informative KV tokens as the KV cache~\cite{he2024zipvl}. Formally, given the current block $c$ and its queries $Q$, keys $K$, and values $V$, we compute the attention score matrix
\begin{equation}
A = \text{Softmax}\Big(\tfrac{QK^\top}{\sqrt{d}} + \textsc{Mask}\Big),
\end{equation}
where the $\textsc{Mask}$ denotes a block-level causal mask.

We then aggregate attention scores across heads and probe queries to form an importance score vector
$\mathbf{m} \in \mathbb{R}^{N_{\text{token}}}$, where each element $\mathbf{m}_i$ measures the relative importance of the $i$-th token.
Subsequently, we select the most informative tokens using a top-$k$ indexing strategy.
Specifically, let $M$ denote the minimal number of tokens whose cumulative importance covers a fraction $\tau$ of the total score.
The retained token indices are given by
\begin{equation}
\mathcal{I}_{\text{keep}} = \text{topk\_index}(\mathbf{m}, M).
\end{equation}
This procedure yields a sparse cache $(K_{\text{sparse}}, V_{\text{sparse}})$ that contains only the most relevant context tokens.

This produces a sparse cache $(K_{\text{sparse}}, V_{\text{sparse}})$ containing only the most relevant context tokens. 

During generation, the sparse KV caches from past blocks are stored in a global KV bank and retrieved based on their semantic similarity with prompt embeddings:
\begin{equation}
\text{sim}_i = \cos\!\big(E_c, E_i\big), \quad i \in \{1, \dots, c{-}1\},
\end{equation}
where $E_c$ is the current prompt's embedding and $\{E_i\}$ are the past ones. The top-$l$ most similar entries are then selected.  
Finally, we concatenate the top-$l$ semantic KV caches (blocks) with the two most recent caches to form the final KV cache:
\begin{align}
({K}^*, {V}^*) &= \textsc{ConcatKV}\Big( \{({K}_j,{V}_j)\}_{j \in \text{seq\_ctx}}, \notag \\
&\quad \{({K}_i, {V}_i)\}_{i \in \text{top-}l} \Big).
\end{align}
where 
$\text{seq\_ctx}=\{c{-}2, c{-}1\}$ (if available). The detailed algorithm is provided in \textit{Appendix: Algorithm}.

Finally, the aggregated KV cache $({K}^*, {V}^*)$ serves as conditional context, combined with the current prompt $y_t$ to guide the generation of the target block:
\begin{equation}
 V_t \sim p_\theta\!\left(\,\cdot \mid {K}^*, {V}^*, y_t \right).   
\end{equation}

\subsection{Block Forcing}
\label{sec:bf}

Although Self Forcing~\cite{huang2025self} serves as an effective strategy through mitigating the training–inference gap, it stabilizes predictions only within a single block and lacks mechanism for maintaining \emph{cross-block} coherence. Moreover, when generating very long videos, a model trained with Self Forcing alone can still lose track of the subject or scene, leading to gradual drift (i.e. the character slowly changing identity or the background progressively melting). 

To this end, we formulate a Block Forcing loss that generalizes the Self Forcing approach to long-range scenarios.
This is achieved by enforcing semantic alignment between the current video block and its most relevant historical context.
Specifically, the top-$l$ past blocks are resampled to match the temporal length of the current block and averaged into a semantic reference $x_{\text{cond}}$, which serves as high-level guidance to maintain long-term coherence.

Formally, the Block Forcing loss penalizes the deviation of the predicted velocity $v_{\text{pred}}$ from both the noise term $\epsilon$ and the semantic reference $x_{\text{cond}}$, weighted by $\gamma \in [0,1]$:
\begin{equation}
\mathcal{L}_{\text{BF}} 
= \mathbb{E}\Big[\| v_{\text{pred}} - (\epsilon - \gamma \cdot x_{\text{cond}}) \|^2\Big].
\end{equation}

This formulation ensures that the model learns not only to denoise the current block correctly but also to remain semantically anchored to the relevant history, thereby reducing temporal drift and improving the stability of long video generation. The total training loss for \textit{interactive long tuning} is $\mathcal{L} = \mathcal{L}_{\text{SF}} + \mathcal{L}_{\text{BF}}$.

\section{InterVBench}

\noindent\textbf{Dataset.}
To tackle the challenge of evaluation for interactive minute-long video generation, we curate a dataset of 1000 videos from diverse open-source sources and annotate them in detail. As shown in Table \ref{tab:datasets}, we collect high-quality video blocks with lengths of \textbf{at least 50 seconds} from DanceTrack~\cite{sun2022dancetrack}, GOT-10k~\cite{huang2019got}, HD-VILA-100M~\cite{xue2022advancing}, and ShareGPT4V~\cite{chen2024sharegpt4v}. To obtain high-quality annotations, we employ GPT-4o as a data engine to generate fine-grained captions for every 2–3 seconds in each video. The detailed prompt can be found in \textit{Appendix: InterVBench}. Human-in-the-loop validation consists of manual visual checks at every stage of data production, including data sourcing, block splitting, and captioning, to ensure high-quality annotations. In the data sourcing stage, human annotators select high-quality videos and determine whether each raw video is suitable for inclusion. In block splitting, human annotators examine samples to verify that each block is free of errors such as incorrect transitions. In captioning, human annotators review the generated descriptions to ensure semantic accuracy and coherence. At each stage, at least two human annotators participate to provide inter-rater reliability.
For metrics, see \textit{Appendix: Metrics}.

\begin{table}[t]
\centering
\caption{\textbf{Overview of the datasets used to construct InterVBench}, where “H” stands for “Human,” “A” stands for “Animal,” and “E” stands for “Environment.”}
\setlength{\tabcolsep}{10pt}
\renewcommand{\arraystretch}{1.2}
\resizebox{0.65\linewidth}{!}{
\begin{tabular}{lcc}
\toprule
\textbf{Dataset} & \textbf{Video Number} & \textbf{Object Classes} \\
\midrule
DanceTrack   & 66  & H (66, 100\%) \\
GOT-10k      & 272 & H (177, 65\%)~ A (54, 20\%)~ E (41, 15\%) \\
HD-VILA-100M & 117 & H (47, 40\%)~ A (35, 30\%)~ E (35, 30\%) \\
ShareGPT4V   & 545 & H (381, 70\%)~ A (82, 15\%)~ E (82, 15\%) \\
\midrule
\textbf{InterVBench}   & \textbf{1000} & \textbf{H (671, 67\%)~ A (171, 17\%)~ E (158, 16\%)} \\
\bottomrule
\end{tabular}
}
\label{tab:datasets}
\end{table}

\begin{table*}[t]
  \centering
  \caption{\textbf{Comparison of different methods on VBench-Long~\cite{huang2024vbench}.}
  We extended VBench-Long to 60 seconds following LongLive~\cite{yang2025longlive}.}
  \label{tab:vbench}
  \setlength{\tabcolsep}{6pt}
  \renewcommand{\arraystretch}{1.2}
  \resizebox{0.9\linewidth}{!}{
    \begin{tabular}{l|c|c|c|c|c|c}
      \toprule
      Method &
      \makecell{Subject \\ Consistency} $\uparrow$ &
      \makecell{Background \\ Consistency} $\uparrow$ &
      \makecell{Motion \\ Smoothness} $\uparrow$ &
      \makecell{Dynamic \\ Degree} $\uparrow$ &
      \makecell{Aesthetic \\ Quality} $\uparrow$ &
      \makecell{Image \\ Quality} $\uparrow$ \\
      \midrule
      MAGI-1 & 0.8320 & 0.8931 & 0.9740 & 0.5537 & 0.5010 & 0.6120 \\
      Self Forcing & 0.8211 & 0.9050 & 0.9799 & 0.6015 & 0.5130 & 0.6218 \\
      PAVDM & 0.8415 & 0.9273 & 0.9769 & 0.6537 & 0.4970 & 0.6280 \\
      FramePack & 0.9019 & 0.9450 & 0.9805 & 0.5715 & 0.5044 & 0.6381 \\
      Self-Forcing++ & 0.9165 & 0.9092 & 0.9803 & 0.5865 & 0.5482 & 0.6453 \\
      Rolling Forcing & 0.9409 & 0.9447 & 0.9865 & 0.6600 & \textbf{0.6350} & 0.6442 \\
      Deep Forcing & 0.9285 & 0.9136 & 0.9819 & 0.7035 & 0.6041 & 0.6455 \\
      StreamDiffusionV2 & 0.9036 & 0.9051 & 0.9745 & 0.4552 & 0.5547 & 0.5528 \\
      LongLive & 0.9403 & 0.9495 & 0.9845 & 0.7321 & 0.5795 & 0.6483 \\
      Infinity-RoPE & 0.9352 & 0.9395 & 0.9710 & 0.5395 & 0.6045 & 0.6475 \\
      SkyReels-V2-DF-1.3B & 0.9391 & 0.9580 & 0.9838 & 0.6529 & 0.5320 & 0.6315 \\
      LCT (MMDiT-3B) & 0.9380 & 0.9623 & 0.9816 & 0.6875 & 0.5200 & 0.6345 \\
      MoC & 0.9398 & \textbf{0.9670} & 0.9851 & 0.7500 & 0.5547 & 0.6396 \\
      \midrule
      \textbf{BIFE-1.3B (Ours)} &
      \textbf{0.9410} & 0.9650 & \textbf{0.9870} &
      \textbf{0.7720} & 0.5839 & \textbf{0.6527} \\
      \bottomrule
    \end{tabular}
  }
\end{table*}

\section{Experiment}

\subsection{Implementation Details}
To ensure a fair comparison, we follow the training strategy in Self-Forcing~\cite{huang2025self} and LongLive~\cite{yang2025longlive}.
We build BIFE upon Wan2.1-T2V-1.3B~\cite{wan2025wan}, which generates 5-second clips (81 frames) at 16 FPS with a resolution of 480p ($832 \times 480$). We first adapt the pretrained model into a 4-step causal-attention student model using a self-forcing DMD pipeline~\cite{huang2025self,yin2024one} on the VidProM dataset~\cite{wang2024vidprom}. We then use the student model and a Wan2.1-T2V-14B teacher model to perform \textit{interactive long tuning} on switch prompts with 60s length. The switch prompts are constructed following LongLive~\cite{yang2025longlive}, where Qwen2-72B-Instruct~\cite{bai2023qwen} generates follow-up prompts conditioned on each original VidProM prompt. During training, each iteration extends the model’s own rollout by generating successive 5s clips until reaching a maximum length of 60s. Each training sample contains exactly one prompt switch, with the switch time uniformly sampled between 5s and 55s. 
The full training process takes approximately 30 hours on 32 NVIDIA H20 GPUs, supported by 192 CPU cores and 960 GB of CPU memory.
During inference, we set each chunk to contain 8 frames, resulting in $8 \times 1560 = 12480$ tokens before sparsification in the semantic sparse KV cache.
We employ AdamW and stepwise decay schedule for all stages of training. The initial learning rate is $1 \times 10^{-4}$, then reduced to $5 \times 10^{-5}$, with the weight decay set to $1 \times 10^{-4}$. For the semantic sparse KV cache, due to the limitation of H20 GPU memory, we use Top-$l$ semantic retrieval with $l=2$.

\subsection{Main Results}

\textbf{Results on VBench-Long.}
We compare our method with state-of-the-art baselines on VBench-Long~\cite{huang2024vbench}. Because the standard VBench-Long protocol is not directly applicable (only 30s), we use the prompts curated by LongLive~\cite{yang2025longlive}, a custom set of 160 interactive 60-second videos, each comprising six successive 10-second prompts. 
As shown in Table~\ref{tab:vbench}, BIFE-1.3B achieves superior performance across the majority of metrics, surpassing both open-source and large-scale proprietary baselines.
Specifically, \texttt{BIFE-1.3B} achieves the best performance in subject consistency (0.9410), motion smoothness (0.9870), dynamic degree (0.7720), and image quality (0.6527), while maintaining competitive results in background consistency (0.9650) and aesthetic quality (0.5839). Compared with prior methods, our model demonstrates strong advantages in both temporal dynamics and perceptual quality. These results highlight the effectiveness of our method in achieving both temporal coherence and high visual quality for long video generation.

\begin{table}[t!]
\centering
\caption{\textbf{Comparison of different methods on InterVBench.} We report InterVBench results on five VDE metrics and five complementary metrics from VBench~\cite{huang2024vbench}. Our \texttt{BIFE} outperforms baselines on the majority.}
\label{tab:intervbench}
\setlength{\tabcolsep}{6pt}
\renewcommand{\arraystretch}{1.3}
\resizebox{0.8\linewidth}{!}{
\begin{tabular}{l|c|c|c|c|c}
\toprule
Method &
\makecell{VDE \\ Subject} $\downarrow$ &
\makecell{VDE \\ Background} $\downarrow$ &
\makecell{VDE \\ Motion} $\downarrow$ &
\makecell{VDE \\ Aesthetic} $\downarrow$ &
\makecell{VDE \\ Clarity} $\downarrow$ \\
\midrule
MAGI-1 & 0.3090 & 0.5000 & 0.0243 & 3.8286 & 2.7225\\
Self Forcing & 0.3716 & 1.6108 & 0.1549 & 3.4683 & 3.0798\\
PAVDM & 1.8292 & 0.9323 & 0.0461 & 2.8957 & 1.9503\\
FramePack & 4.3984 & 5.9421 & 0.0387 & 1.4751 & 4.2513 \\
SkyReels-V2-DF-1.3B & 0.1085 & 0.3179 & 0.0195 & 1.2083 & 0.9365 \\
\midrule
\textbf{BIFE-1.3B (Ours)} & \textbf{0.0844} & \textbf{0.2945} & \textbf{0.0119} & \textbf{0.9618} & \textbf{0.7551}\\
\midrule
Method &
\makecell{Subject \\ Consistency} $\uparrow$ &
\makecell{Background \\ Consistency} $\uparrow$ &
\makecell{Motion \\ Smoothness} $\uparrow$ &
\makecell{Aesthetic \\ Quality} $\uparrow$ &
\makecell{Image \\ Quality} $\uparrow$ \\
\midrule
MAGI-1 & 0.8992 & 0.9078 & 0.9947 & \textbf{0.6508} & 0.6662\\
Self Forcing & 0.8481 & 0.8203 & 0.9947 & 0.6283 & 0.6805\\
PAVDM & 0.8640 & 0.8924 & 0.9926 & 0.5267 & 0.6567\\
FramePack & 0.9001 & 0.8791 & 0.9949 & 0.6043 & \textbf{0.6972} \\
SkyReels-V2-DF-1.3B & 0.9418 & 0.9579 & 0.9931 & 0.6035 & 0.6835 \\
\midrule
\textbf{BIFE-1.3B (Ours)} & \textbf{0.9597} & \textbf{0.9588} & \textbf{0.9956} & 0.6047 & 0.6852\\
\bottomrule
\end{tabular}
}
\end{table}

\begin{table*}[t]
  \centering
    \caption{\textbf{Efficiency comparison under 60-second video generation.}}
    \label{tab:efficiency}
    \vskip -0.05in
    \setlength{\tabcolsep}{6pt}
    \renewcommand{\arraystretch}{1.2}
    \resizebox{0.65\linewidth}{!}{
      \begin{tabular}{l|c|c|c|c}
        \toprule
        Method &
        \makecell{Latency \\ $\downarrow$ (s)} &
        \makecell{Throughput \\ $\uparrow$ (FPS)} &
        \makecell{GPU Memory \\ $\downarrow$ (GB)} &
        \makecell{KV Cache Size \\ $\downarrow$} \\
        \midrule
        Self Forcing & 0.78 & 15.38 & 40 & 49920 \\
        LongLive & 0.76 & 20.70 & 45 & 49920 \\
        \rowcolor{yellow!30} \textbf{BIFE (Ours, $\tau=0.9$)} &
        \textbf{0.75} & \textbf{21.10} & \textbf{32} & \textbf{28950} \\
        \bottomrule
      \end{tabular}
    }
\end{table*}

\noindent\textbf{Results on InterVBench.} We first compare our method with several open-source long video generation baselines on InterVBench, including MAGI-1~\cite{teng2025magi}, Self Forcing~\cite{huang2025self}, PAVDM~\cite{xie2025progressive}, FramePack~\cite{zhang2025packing}, and SkyReels-V2-DF-1.3B~\cite{chen2025skyreels}. As shown in Table~\ref{tab:intervbench}, our BIFE-1.3B consistently outperforms these methods across most VDE metrics and complementary metrics from VBench. In particular, \texttt{BIFE} achieves the lowest error scores on all five VDE metrics, reducing subject drift, background inconsistency, motion degradation, and perceptual losses compared to strong baselines such as SkyReels-V2-DF-1.3B. On complementary VBench metrics, \texttt{BIFE} also delivers the highest subject consistency (0.9597) and background consistency (0.9588), as well as superior motion smoothness (0.9956). Although \texttt{BIFE} does not achieve the best score on aesthetic quality, it maintains competitive performance in this dimension while delivering state-of-the-art results overall across both VDE and VBench consistency metrics. These results demonstrate that our method not only improves long-term coherence but also balances fidelity and aesthetics in long video generation.

\noindent\textbf{Efficiency.}
To showcase the efficiency of our method in long video generation, we compare against Self Forcing~\cite{huang2025self} and LongLive~\cite{yang2025longlive} under a 60s video generation with a 4-step distillation setting, as shown in Table~\ref{tab:efficiency}. For fair comparison, we follow the original LongLive setting with a KV cache size of 49{,}920 tokens. Our method adopts a semantic sparse KV cache with default $\tau = 0.9$, which retains only 56\% of the original KV cache size. This design leads to lower latency, higher throughput, and reduced GPU memory usage compared to all baselines.
Moreover, we analyze the overhead of sub-stages in terms of end-to-end computation time: the semantic retrieval overhead accounts for 25.66\%, and the KV cache sparsification overhead accounts for 31.00\% of the total computation time.

\begin{figure*}[t]
\centering

\begin{subfigure}[t]{0.24\textwidth}
    \includegraphics[width=\linewidth]{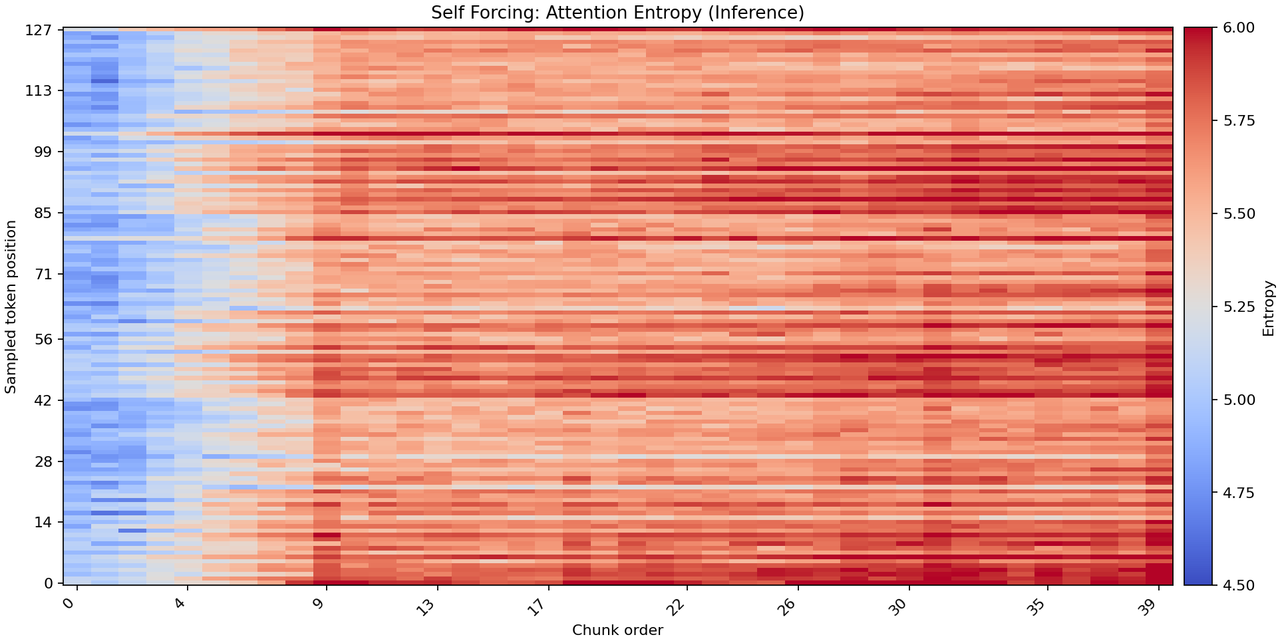}
    \caption{Inference-time (Self Forcing)}
\end{subfigure}
\hfill
\begin{subfigure}[t]{0.24\textwidth}
    \includegraphics[width=\linewidth]{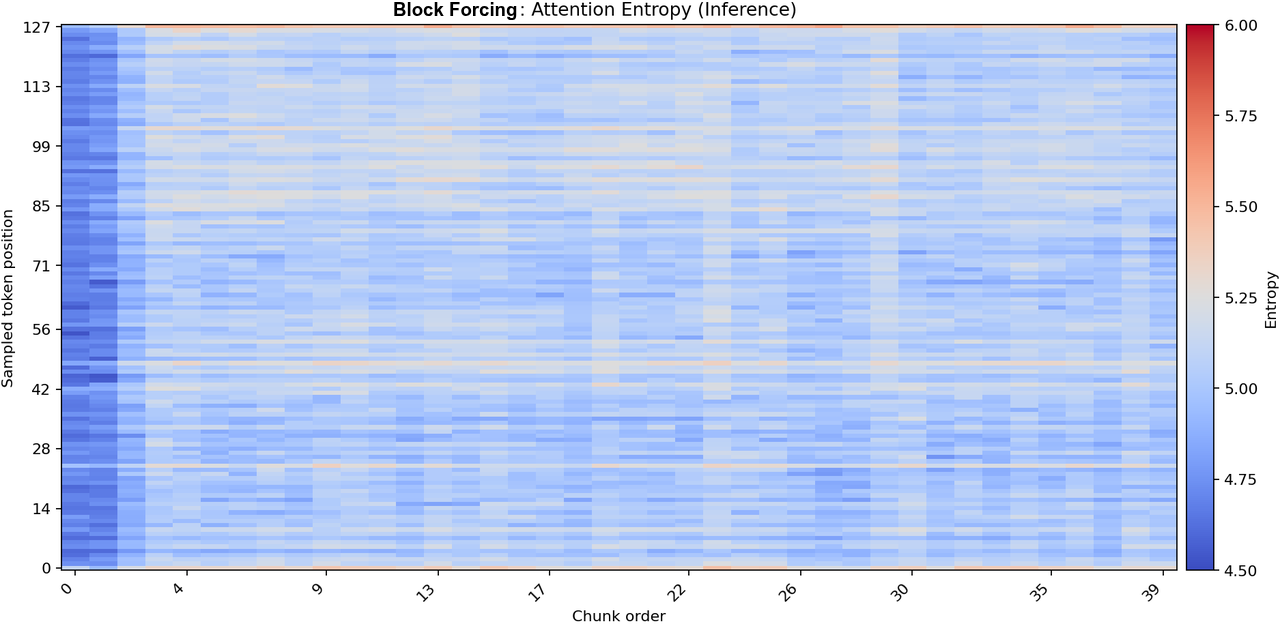}
    \caption{Inference-time (Block Forcing)}
\end{subfigure}
\hfill
\begin{subfigure}[t]{0.24\textwidth}
    \includegraphics[width=\linewidth]{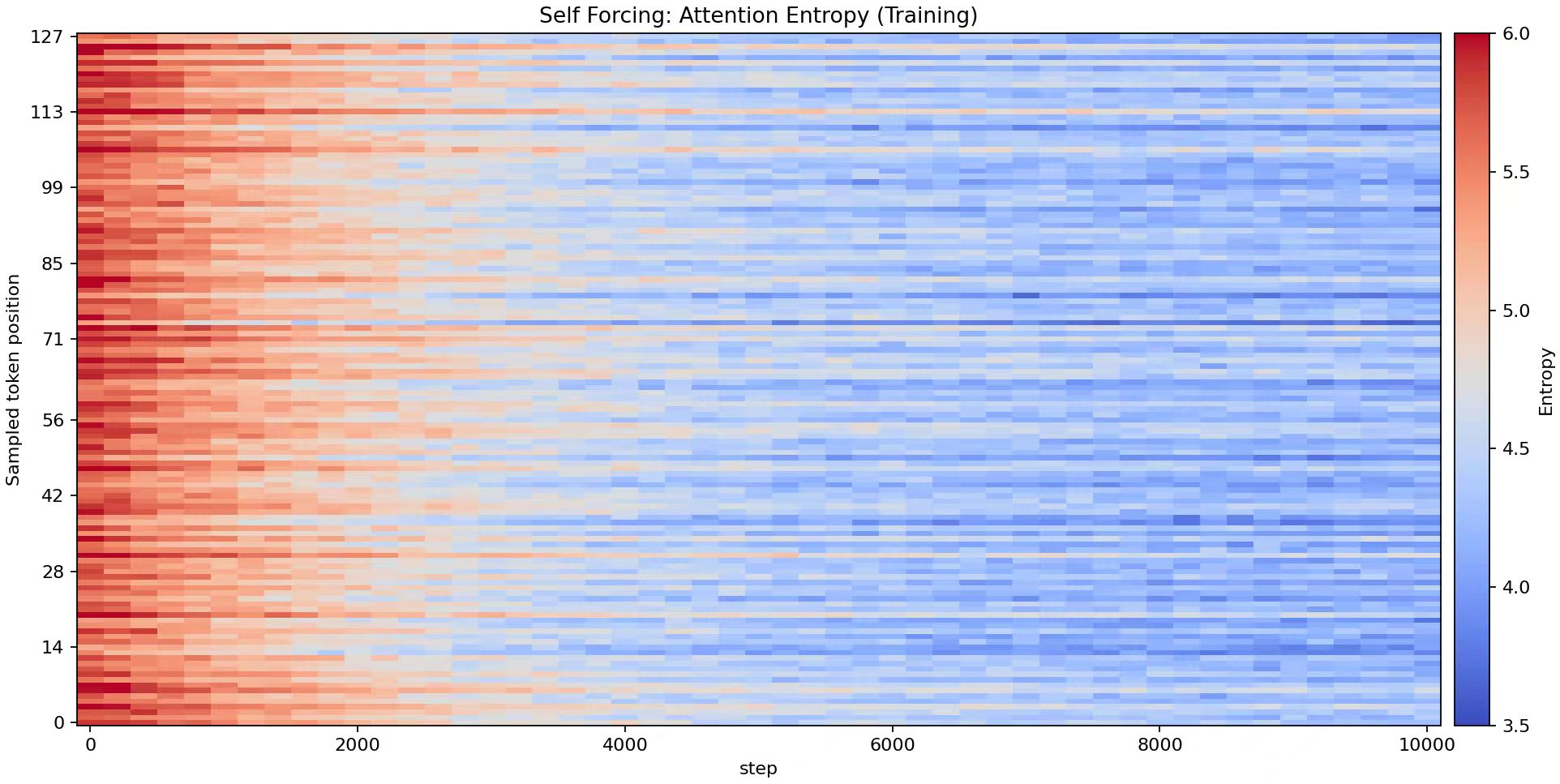}
    \caption{Training-time (Self Forcing)}
\end{subfigure}
\hfill
\begin{subfigure}[t]{0.24\textwidth}
    \includegraphics[width=\linewidth]{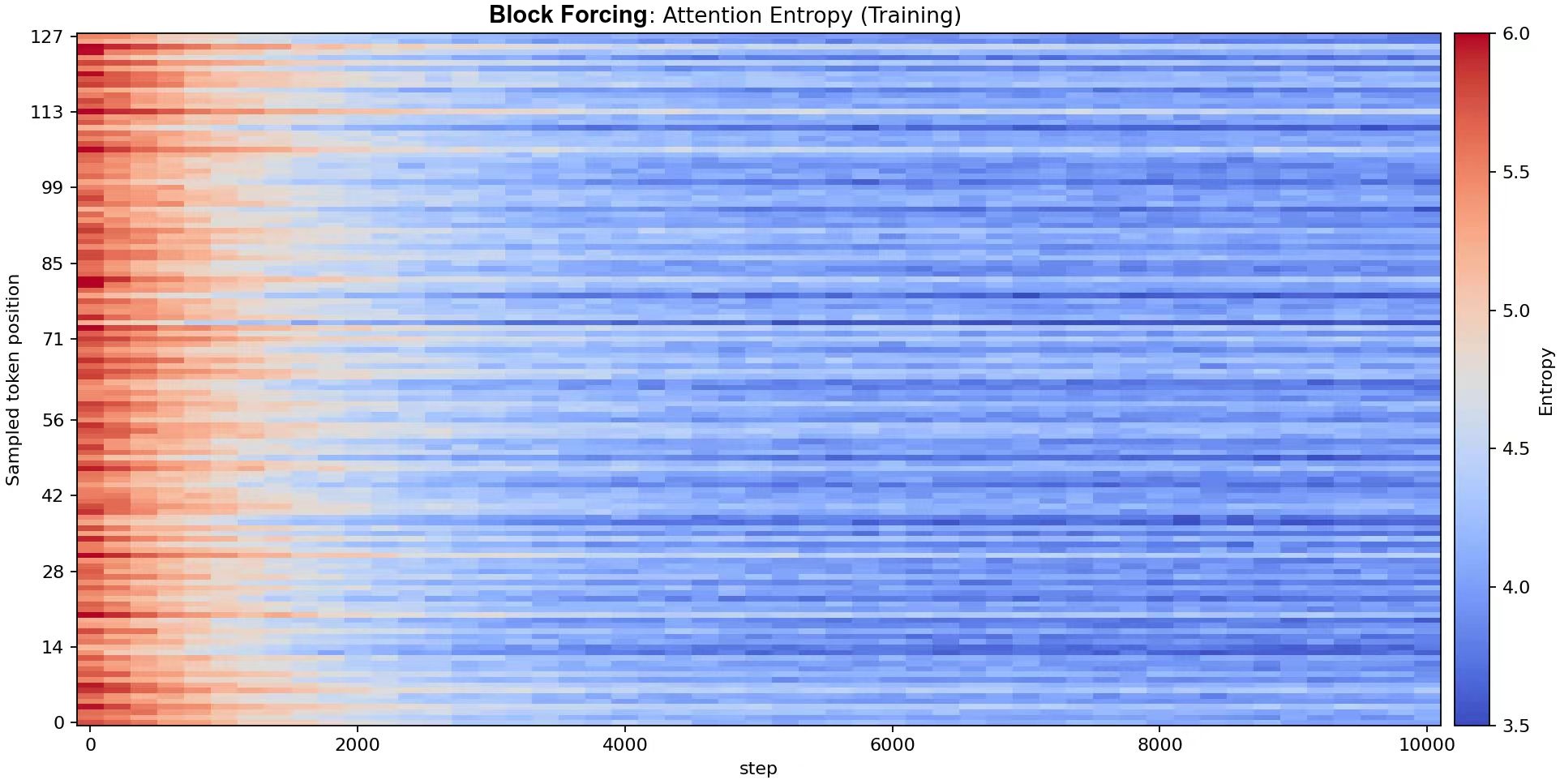}
    \caption{Training-time (Block Forcing)}
\end{subfigure}

\caption{
\textbf{Token entropy comparison.}
The y-axis denotes the sampled token position and the x-axis represents the inference chunk order or training step.
}
\label{fig:entropy}
\end{figure*}

\textbf{Visualization.}
More visualization results of BIFE can be found in Figs.~\ref{fig:more-1},~\ref{fig:more-2},~\ref{fig:more-3},~\ref{fig:more-4},~\ref{fig:more-5},~\ref{fig:more-6}, and in the \textit{Appendix}.

\subsection{Ablation Study}

\begin{table*}[t]
    \centering
    \caption{\textbf{Ablation on Block Forcing, Self Forcing, and $\gamma$, which controls the strength of Block Forcing.}}
    \label{tab:forcing}
    \vskip -0.05in
    \setlength{\tabcolsep}{6pt}
    \renewcommand{\arraystretch}{1.2}
    \resizebox{0.7\linewidth}{!}{
      \begin{tabular}{l|c|c|c|c|c}
        \toprule
        Method &
        \makecell{VDE \\ Subject} $\downarrow$ &
        \makecell{VDE \\ Background} $\downarrow$ &
        \makecell{VDE \\ Motion} $\downarrow$ &
        \makecell{VDE \\ Aesthetic} $\downarrow$ &
        \makecell{VDE \\ Clarity} $\downarrow$ \\
        \midrule
        Self Forcing & 0.0885 & 0.3155 & 0.0169 & 0.9658 & 0.7630 \\
        Block Forcing & 0.0861 & 0.3015 & 0.0137 & 0.9673 & 0.7618 \\
        \rowcolor{yellow!30} \textbf{SF + BF} &
        \textbf{0.0844} & \textbf{0.2945} & \textbf{0.0119} &
        \textbf{0.9618} & \textbf{0.7551} \\
        \midrule
        $\gamma = 0.3$ & 0.0852 & 0.2971 & 0.0126 & 0.9630 & 0.7568 \\
        \rowcolor{yellow!30} $\gamma = 0.5$ &
        \textbf{0.0844} & 0.2945 & \textbf{0.0119} &
        \textbf{0.9618} & \textbf{0.7551} \\
        $\gamma = 0.7$ & 0.0850 & \textbf{0.2943} & 0.0123 & 0.9627 & 0.7560 \\
        \bottomrule
      \end{tabular}
    }
\end{table*}

\noindent\textbf{Block Forcing.}
As shown in Table~\ref{tab:forcing}, existing data-free strategies such as Self Forcing~\cite{huang2025self} alleviate short-term exposure bias but remain insufficient for long-horizon block diffusion.
In contrast, Block Forcing significantly improves performance across all VDE metrics and achieves the highest consistency scores on InterVBench. By explicitly modeling inter-block dependencies, Block Forcing directly targets interactive long-horizon generation.

Moreover, $\gamma$ controls the strength of Block Forcing, balancing model prediction and KV-based supervision. Larger $\gamma$ enforces stronger alignment with history, while smaller $\gamma$ allows more flexibility. In Table~\ref{tab:forcing}, $\gamma$ is set empirically, and $\gamma = 0.5$ achieves a good balance between consistency and generation diversity.

To analyze the effect of Block Forcing at the token level, we visualize token entropy and compare it with Self Forcing. Inference-time results are shown in Fig.~\ref{fig:entropy}(a,b), and training-time results in Fig.~\ref{fig:entropy}(c,d).
At inference, token entropy in Self Forcing increases with chunk progression and becomes significantly higher than that of Block Forcing, indicating more severe error accumulation. In contrast, Block Forcing maintains lower entropy, leading to more stable long-horizon generation.
During training, token entropy in Block Forcing decreases faster than in Self Forcing, suggesting more stable optimization and faster convergence.

\begin{table}[t]
\centering
\caption{Ablation on KV cache settings.}
\label{tab:kv}
\vskip -0.05in
\setlength{\tabcolsep}{6pt}
\renewcommand{\arraystretch}{1.5}
\resizebox{0.8\linewidth}{!}{
\begin{tabular}{l|c|c|c|c|c}
\toprule
Method &
\makecell{VDE \\ Subject} $\downarrow$ &
\makecell{VDE \\ Background} $\downarrow$ &
\makecell{VDE \\ Motion} $\downarrow$ &
\makecell{VDE \\ Aesthetic} $\downarrow$ &
\makecell{VDE \\ Clarity} $\downarrow$ \\
\midrule
Rolling KV & 0.0961 & 0.3519 & 0.0547 & 0.9815 & 0.7913\\
Dynamic Sparse KV ($\tau = 0.97$) & 0.0927 & 0.3074 & 0.0253 & 0.9781 & 0.7730\\
Dynamic Sparse KV ($\tau = 0.98$) & 0.0910 & 0.3040 & 0.0239 & 0.9716 & 0.7652\\
\textbf{Semantic Sparse KV} ($\tau = 0.97$) & 0.0869 & 0.2988 & 0.0153 & 0.9684 & 0.7570 \\
\rowcolor{yellow!30} \textbf{Semantic Sparse KV} ($\tau = 0.98$) &
0.0844 & \textbf{0.2945} & 0.0119 &
\textbf{0.9618} & \textbf{0.7551}\\
\textbf{Semantic Sparse KV} ($\tau = 0.99$) &
\textbf{0.0841} & 0.2960 & \textbf{0.0114} & 0.9635 & 0.7563\\
\bottomrule
\end{tabular}
}
\end{table}

\noindent\textbf{KV cache.}
We further explore rolling KV~\cite{huang2025self}, dynamic sparse KV~\cite{he2024zipvl}, and our semantic sparse KV under different attention thresholds $\tau$. As shown in Table~\ref{tab:kv}, our semantic sparse KV cache with $\tau=0.98$ achieves the best trade-off. When $\tau > 0.98$, results change slightly, with some metrics improving while others degrade.

\noindent\textbf{Extra backbones.}
We include an additional backbone, CogVideoX-2B 480p~\cite{yang2024cogvideox}, and compare it with two KV cache paradigms: Self Forcing~\cite{huang2025self} (sliding window) and LongLive~\cite{yang2025longlive} (sliding window with re-cache), as shown in Table~\ref{tab:backbone}. All methods are evaluated on VBench-Long extended to 60 seconds.
Despite CogVideoX-2B models having lower overall performance than Wan 2.1 1.3B, BIFE consistently outperforms both baselines under the same setting, demonstrating strong generality and robustness across different backbones and long-video generation paradigms.

\section{Human Evaluation for InterVBench}

To ensure that InterVBench aligns closely with human judgment across all evaluation dimensions, we invite 25 human experts to conduct preference labeling on the test set of generated videos, following the protocol of VBench~\cite{huang2024vbench}. Specifically, we demonstrate the alignment between InterVBench scores and human annotations.
Given human annotations, we compute the win ratio for each model based on pairwise comparisons. For each pair, if a model's video is preferred, it receives a score of 1, while the other receives 0; in case of a tie, both receive 0.5. The win ratio is then calculated as the total score divided by the number of pairwise comparisons.
The Table~\ref{tab:intervbench_alignment} reflects strong alignment between InterVBench and human evaluations. 

\begin{table}[t]
\centering
\caption{Extra backbones with CogVideoX-2B 480P~\cite{yang2024cogvideox} on VBench-Long extended to 60 seconds~\cite{yang2025longlive}.}
\label{tab:backbone}
\vskip -0.05in
\setlength{\tabcolsep}{6pt}
\renewcommand{\arraystretch}{1.2}
\resizebox{\linewidth}{!}{
\begin{tabular}{l|c|c|c|c|c|c}
\toprule
Method &
\makecell{Subject \\ Consistency} $\uparrow$ &
\makecell{Background \\ Consistency} $\uparrow$ &
\makecell{Motion \\ Smoothness} $\uparrow$ &
\makecell{Dynamic \\ Degree} $\uparrow$ &
\makecell{Aesthetic \\ Quality} $\uparrow$ &
\makecell{Image \\ Quality} $\uparrow$ \\
\midrule
Self Forcing (CogVideoX-2B) & 0.8136 & 0.8091 & 0.8254 & 0.3365 & 0.4750 & 0.5523\\
LongLive (CogVideoX-2B) & 0.8442 & 0.8119 & 0.8375 & 0.4890 & 0.4795 & 0.5780\\
\textbf{BIFE (CogVideoX-2B)} & \textbf{0.8651} & \textbf{0.8359} & \textbf{0.8562} & \textbf{0.5021} & \textbf{0.4881} & \textbf{0.5805}\\
BIFE (Wan 2.1 5B) & 0.9410 & 0.9650 & 0.9870 & 0.7720 & 0.5839 & 0.6527\\
\bottomrule
\end{tabular}
}
\end{table}

\begin{table}[t!]
\centering
\footnotesize
\setlength{\tabcolsep}{4pt}
\caption{\textbf{The alignment between InterVBench scores and human annotations.} 
For each evaluation dimension and each video generation model, we report results in the format: 
``InterVBench Win Ratio (left) / Human Win Ratio (right)''.}
\label{tab:intervbench_alignment}
\resizebox{0.8\linewidth}{!}{
\begin{tabular}{lccccc}
\toprule
\textbf{Method} & \textbf{Subject} & \textbf{Background} & \textbf{Motion} & \textbf{Aesthetic} & \textbf{Clarity} \\
\midrule
MAGI-1 & 41.5 / 42.3 & 44.2 / 43.1 & 40.8 / 41.6 & 46.9 / 45.7 & 43.7 / 44.5 \\
Self Forcing & 48.6 / 47.2 & 52.1 / 50.8 & 47.5 / 46.1 & 49.3 / 50.0 & 45.2 / 46.3 \\
PAVDM & 53.4 / 52.0 & 50.2 / 49.1 & 55.1 / 53.8 & 48.6 / 47.9 & 52.7 / 51.5 \\
FramePack & 56.8 / 55.4 & 54.9 / 53.6 & 52.0 / 50.9 & 57.2 / 56.1 & 49.8 / 48.6 \\
SkyReels-V2-DF-1.3B & 58.1 / 56.9 & 57.5 / 56.2 & 41.9 / 43.0 & 56.4 / 55.1 & 54.2 / 53.0 \\
\textbf{BIFE (Ours)} & \textbf{59.2 / 58.0} & \textbf{58.7 / 57.4} & \textbf{56.5 / 55.2} & \textbf{55.8 / 56.3} & \textbf{57.1 / 55.9} \\
\bottomrule
\end{tabular}
}
\end{table}

\section{Conclusion}

We present BIFE, an autoregressive diffusion framework for interactive minute-long video generation that explicitly addresses long-horizon error accumulation and interaction consistency. By introducing a semantic sparse KV cache, our method preserves long-range dependencies through retrieval-based conditioning, overcoming the limitations of sliding-window KV designs. In addition, the proposed Block Forcing objective which effectively reduces cross-block drift and stabilizes long-term generation.
To support the evaluation, we introduce InterVBench, a benchmark with fine-grained annotations and the proposed Video Drift Error (VDE) metrics, enabling systematic assessment of long-horizon consistency and interaction quality. Extensive experiments on both InterVBench and VBench demonstrate that BIFE achieves state-of-the-art performance while maintaining strong efficiency.
For \textit{limitation and future work}, see \textit{Appendix}.

\bibliographystyle{plain}
\bibliography{paper}

\clearpage
\appendix

\section{Algorithm: Semantic Sparse KV Cache}
\label{app:algo}

See Algorithm~\ref{alg:semantic_sparse_kv_bank_final} and~\ref{alg:build_sparse_kv}.

\begin{algorithm}
\caption{Semantic Sparse KV Cache}
\label{alg:semantic_sparse_kv_bank_final}
\begin{algorithmic}

\Require blocks $\{X_i\}_{i=1}^{N}$, prompts $\{\mathcal{Y}_i\}_{i=1}^{N}$,
target $t{=}N$, threshold $\tau$, top-$K$, drop $p_{\text{drop}}$
\Ensure final KV cache $(\mathsf{K}^{*}, \mathsf{V}^{*})$ for $X_t$

\State $\text{KV\_BANK} \gets \varnothing$
\Comment{dictionary: $i \mapsto (\mathsf{K}^{(i)}_{\text{sparse}}, \mathsf{V}^{(i)}_{\text{sparse}})$}

\For{$c \in \{1,\dots,N{-}1\}$}
  \If{$c \notin \text{KV\_BANK}$}
    \State $(\mathsf{K}^{(c)}_{\text{sparse}}, \mathsf{V}^{(c)}_{\text{sparse}}) \gets
    \textsc{BuildSparseKV}(X_c,\mathcal{Y}_c,\tau)$
    \State $\text{KV\_BANK}[c] \gets
    (\mathsf{K}^{(c)}_{\text{sparse}}, \mathsf{V}^{(c)}_{\text{sparse}})$
  \EndIf
\EndFor

\State $\text{seq\_ctx} \gets \{N{-}3, N{-}2\}$
\State $E_t \gets \textsc{MeanEmbed}(\mathcal{Y}_t)$
\State $\mathcal{S} \gets \{1,\dots,N{-}1\} \setminus \text{seq\_ctx}$

\For{$i \in \mathcal{S}$}
  \State $E_i \gets \textsc{T5-Embed}(\mathcal{Y}_i)$
  \State $sim_i \gets \cos(E_t, E_i)$
\EndFor

\State $\text{TopKIdx} \gets
\text{argsort}(\{sim_i\}_{i\in\mathcal{S}})[-K:]$

\State $(\mathsf{K}_{\text{seq}}, \mathsf{V}_{\text{seq}}) \gets
\textsc{ConcatKV}(\{\text{KV\_BANK}[j] : j \in \text{seq\_ctx}\})$

\State $(\mathsf{K}_{\text{sem}}, \mathsf{V}_{\text{sem}}) \gets
\textsc{ConcatKV}(\{\text{KV\_BANK}[i] : i \in \text{TopKIdx}\})$

\State $(\mathsf{K}^{*}, \mathsf{V}^{*}) \gets
\textsc{ConcatKV}((\mathsf{K}_{\text{seq}}, \mathsf{V}_{\text{seq}}),
(\mathsf{K}_{\text{sem}}, \mathsf{V}_{\text{sem}}))$

\State \textbf{return} $(\mathsf{K}^{*}, \mathsf{V}^{*})$

\end{algorithmic}
\end{algorithm}

\begin{algorithm}
\caption{BuildSparseKV: Dynamic Sparse KV Cache}
\label{alg:build_sparse_kv}
\begin{algorithmic}
\State \Comment{BuildSparseKV($X, \mathcal{Y}, \tau$)}
\State ${H} \gets \textsc{Encode}(X,\mathcal{Y})$
\State \Comment{model input states}
\State $Q, K, V \gets \textsc{Project}(H)$
\State $(Q, K) \gets \textsc{RoPE}(Q, K)$
\State $q_{\text{len}} \gets \text{length}(Q)$
\If{$q_{\text{len}} > 1$}
\State \Comment{prefill stage}
\State $\mathcal{I}_{\text{probe}} \gets \textsc{Concat}(\text{Recent}(64), \text{Random}(64))$
\State $Q_{\text{probe}} \gets Q[:, \mathcal{I}_{\text{probe}}, :]$
\State ${A} \gets \textsc{Softmax}\left(\frac{Q_{\text{probe}} K^\top}{\sqrt{d}} + \textsc{CausalMask}\right)$
\State ${s} \gets \sum_{\text{heads, probe}} A$
\State ${m} \gets \textsc{CumMean}(s)$
\State $M \gets \textsc{CoverCount}(m, \tau)$
\State $\mathcal{I}_{\text{keep}} \gets \textsc{Top-K}(m, M)$
\State \textbf{return} $(K[:, \mathcal{I}_{\text{keep}}, :], V[:, \mathcal{I}_{\text{keep}}, :])$
\Else
\State \textbf{return} $(K, V)$
\EndIf
\end{algorithmic}
\end{algorithm}

\section{InterVBench}

\subsection{Prompts for InterVBench's Data Engine}
\label{app:prompts}

\begin{center}
\begin{tcolorbox}[promptbox, before skip=0pt, after skip=0pt]
\textbf{Role.} Act as a professional video content analyst. Describe a given video frame in English.\\
\textbf{Context.} The previous frame was described as: \emph{"\{previous\_description\}"}. Use this as context to ensure temporal coherence.\\
\textbf{Instruction.} Write a single, descriptive paragraph that: 
\begin{itemize}[itemsep=1pt,leftmargin=10pt]
\item Identifies the \textbf{main subject}, their specific actions, and expressions. 
\item Describes the \textbf{environment and background}, including setting and lighting. 
\item Highlights the \textbf{cinematic quality}, such as composition, color palette, and atmosphere (tense, serene, spectacular). 
\end{itemize}
\textbf{Constraints.} Output must be \textbf{one coherent paragraph}, written in natural language prose, without bullet points or numbered lists.\\
\textbf{Return.} The paragraph description of the current frame.
\end{tcolorbox}
\end{center}

\section{Metrics.}
\label{sec:metrics}
Drift penalties have been widely adopted to address information dilution~\cite{li2025longdiff} and degradation~\cite{lu2024freelong} in long video generation. For example, IP-FVR~\cite{han2025show} focuses on preserving identity consistency, while MoCA~\cite{xie2025moca} employs an identity perceptual loss to penalize frame-to-frame identity drift. 
Inspired by the commonly used metrics MAPE and WMAPE~\cite{kim2016new,de2016mean}, we propose a new metric called Video Drift Error (VDE) to measure changes in video quality.
We further design 5 long video generation metrics based on VDE. The core idea involves dividing a long video into multiple segments, each evaluated according to specific quality
metrics (clarity, motion smoothness, etc). Specifically, (1) \textit{VDE Clarity} measures temporal drift in image sharpness, where creeping blur
increases the score, while a low value indicates stable clarity over time. (2) \textit{VDE Motion} measures drift in motion smoothness, where a low score
indicates consistent dynamics without jitter or freezing. (3) \textit{VDE Aesthetic} measures drift in visual appeal, where a low score indicates sustained
and coherent aesthetics over time. (4) \textit{VDE Background} measures background stability, where a low score indicates a consistent setting without
drift or flicker over time. (5) \textit{VDE Subject} tracks identity drift, where a low score indicates the subject remains consistently recognizable over time. Following previous works~\cite{guo2025long,cai2025mixture}, we also include five complementary metrics from VBench~\cite{huang2024vbench}. The details are included in \textit{Appendix: InterVBench}.

\subsection{InterVBench Metrics}
\label{app:metrics}

\subsubsection{Preliminaries: Mean Absolute Percentage Error}

Mean Absolute Percentage Error (MAPE) and Weighted Mean Absolute Percentage Error (WMAPE) are widely adopted evaluation metrics in forecasting \citep{kim2016new}, time series analysis \citep{de2016mean}, and increasingly in video quality assessment tasks \citep{huang2020quality}. 
MAPE measures the average relative deviation between predicted values $\hat{y}_i$ and ground-truth values $y_i$, expressed as a percentage:  

\begin{equation}
\text{MAPE} = \frac{100}{N} \sum_{i=1}^N \left| \frac{y_i - \hat{y}_i}{y_i} \right|.
\end{equation}

Although simple and interpretable, MAPE can be biased when actual values $y_i$ are close to zero. 
To address this issue, WMAPE normalizes the absolute error by the sum of actual values, making the metric scale-invariant and more robust in practice:  
\begin{equation}
\text{WMAPE} = \frac{\sum_{i=1}^N |y_i - \hat{y}_i|}{\sum_{i=1}^N |y_i|}.
\end{equation}

These metrics provide interpretable percentage-based measures of consistency and prediction accuracy, 
and can be directly applied to quantify deviations across frames or segments in video tasks \citep{huang2020quality}.

\subsubsection{Video Drift Error (VDE)}

Inspired by the WMAPE \citep{kim2016new,de2016mean}, we propose a new metric called Video Drift Error (VDE) to measure changes in video quality. The core idea involves dividing a long video into multiple smaller segments, each evaluated according to specific quality metrics (such as clarity, motion smoothness, etc). These scores are then used to calculate the relative change compared to the first segment. For long video generation, small quality deviations may accumulate within each short time segment. Over time, these deviations gradually build up \citep{li2025longdiff,lu2024freelong}. This accumulation error can be quantified and detected through VDE. Specifically, a high VDE value indicates significant fluctuations or degradation in video quality as playback progresses, while a low VDE value suggests consistent quality levels throughout.
Similar drift penalties have been introduced in works such as IP-FVR \citep{han2025show}, which focuses on preserving identity consistency, and MoCA \citep{xie2025moca}, which employs an identity perceptual loss to penalize frame-to-frame identity drift.
Therefore, monitoring VDE during long-term video generation helps identify potential quality degradation trends and allows timely corrective actions to be taken.

Specifically, the method first divides the video into \( N \) smaller segments of equal duration: 
$ V = \{ S_1, S_2, \dots, S_N \},$ 
where \( V \) is the full video, and \( S_i \) represents the \( i \)-th segment.  

Then the method evaluate each segment by applying a quality evaluation function (e.g., \( \text{metric\_function} \)) to compute a score \( Q_i \) for each segment \( S_i \): 
\begin{equation}
 Q_i = \text{metric\_function}(S_i), \quad \forall i \in \{1, 2, \dots, N\}. 
\end{equation}
Furthermore, the method compute rate of change which 
calculates the relative change \( \Delta_i \) in quality scores from the first segment (\( Q_1 \)) for all subsequent segments (\( i \geq 2 \)): 
\begin{equation}
 \Delta_i = \frac{Q_i - Q_1}{Q_1}.  
\end{equation}
The final VDE value is derived as a weighted sum of absolute rate changes, using linear or logarithmic weights \( w_i \):  
\begin{equation}
\text{VDE} = \sum_{i=2}^N w_i \cdot |\Delta_i|.
\end{equation}

\subsubsection{VDE Metrics}

\paragraph{Metric-specific VDEs.}
Given the VDE shell defined in the preliminaries (reference block $S_1$, per-block scores $m_i$, and weights $w_i$), each metric instantiates $m_i$ as follows; the VDE value is then
\begin{equation}
\mathrm{VDE}_{(\cdot)} \;=\; \sum_{i=2}^{N} w_i \,\frac{\lvert m_i - m_1\rvert}{m_1},
\quad
w_i \in \big\{\,N-i+1,\ \log(N-i+1)\,\big\}.
\end{equation}

\paragraph{VDE Clarity ($\downarrow$).}

It evaluates temporal drift in image sharpness (defocus/blur). For long videos, creeping blur or inconsistent deblurring raises $\mathrm{VDE}_{\text{clar}}$, while a low value indicates stable perceived clarity over time.

Let $f_t\in S_i$ be frames and $Y_t$ their luminance. Define per-frame sharpness by Laplacian variance and average within the block:
\begin{align}
m_i^{\text{clar}} \;=\; \frac{1}{|S_i|}\sum_{t\in S_i}\operatorname{Var}\!\big(\nabla^2 Y_t\big),\\
\mathrm{VDE}_{\text{clar}} \;=\; \sum_{i=2}^{N} w_i \frac{\lvert m_i^{\text{clar}}-m_1^{\text{clar}}\rvert}{m_1^{\text{clar}}}.
\end{align}

\paragraph{VDE Motion ($\downarrow$).}

It tracks drift in motion magnitude/smoothness (pace and jitter). Long-sequence generators often change kinetic behavior over time; a low $\mathrm{VDE}_{\text{mot}}$ signals consistent dynamics without late-stage jitter or freezing.

Let ${u}_t$ denote the optical flow between consecutive frames, and define the per-frame motion energy as $E({u}_t)=\|{u}_t\|_2$. 
Alternatively, one may compute a motion-smoothness score $s_t$ based on inter-frame differences. 
The block-level score is then
\begin{equation}
m_i^{\text{mot}} \;=\; \frac{1}{|S_i|-1}\sum_{t\in S_i} E({u}_t)
\quad \text{or} \quad
m_i^{\text{mot}} \;=\; \frac{1}{|S_i|}\sum_{t\in S_i} s_t,
\end{equation}
and the final penalty is
\begin{equation}
\mathrm{VDE}_{\text{mot}} \;=\; \sum_{i=2}^{N} w_i \frac{\lvert m_i^{\text{mot}}-m_1^{\text{mot}}\rvert}{m_1^{\text{mot}}}.
\end{equation}

\paragraph{VDE Aesthetic ($\downarrow$).}

It measures drift in global visual appeal (composition, color harmony, lighting). In long videos, style can drift or collapse; low $\mathrm{VDE}_{\text{aes}}$ indicates sustained, coherent aesthetics along the timeline.

Let $A(f_t)$ be a learned aesthetic predictor applied per frame; average within each block:
\begin{align}
m_i^{\text{aes}} \;=\; \frac{1}{|S_i|}\sum_{t\in S_i} A(f_t),
\\
\mathrm{VDE}_{\text{aes}} \;=\; \sum_{i=2}^{N} w_i \frac{\lvert m_i^{\text{aes}}-m_1^{\text{aes}}\rvert}{m_1^{\text{aes}}}.
\end{align}

\paragraph{VDE Background ($\downarrow$).}

It evaluates stability/consistency of the background (camera drift, flicker, texture boil). Long videos often accumulate spurious background motion; low $\mathrm{VDE}_{\text{bg}}$ reflects a stable setting that does not “melt” over time.

Let $\mathbb{B}_t$ be a background mask and ${u}_t(x)$ the flow at pixel $x$. Define per-frame background staticness $\phi_t=\frac{1}{|\mathbb{B}_t|}\sum_{x\in\mathbb{B}_t}{1}\big(\|{u}_t(x)\|\le\tau\big)$ and average per block:
\begin{equation}
m_i^{\text{bg}} \;=\; \frac{1}{|S_i|}\sum_{t\in S_i}\phi_t,
\qquad
\mathrm{VDE}_{\text{bg}} \;=\; \sum_{i=2}^{N} w_i \frac{\lvert m_i^{\text{bg}}-m_1^{\text{bg}}\rvert}{m_1^{\text{bg}}}.
\end{equation}

\paragraph{VDE Subject ($\downarrow$).}

It captures drift in subject identity/attributes (face morphing, color/outfit changes). For long generations, identity can subtly shift; low $\mathrm{VDE}_{\text{subj}}$ indicates the protagonist remains recognizably consistent throughout.

Let $E(\cdot)$ be a subject-identity encoder and $\bar{{e}}_1$ the mean embedding over subject crops in $S_1$. Define per-frame identity similarity $s_t=\cos\!\big(E(\text{crop}_t),\bar{{e}}_1\big)$ and average within the block:
\begin{equation}
m_i^{\text{subj}} \;=\; \frac{1}{|S_i|}\sum_{t\in S_i} s_t,
\quad
\mathrm{VDE}_{\text{subj}} \;=\; \sum_{i=2}^{N} w_i \frac{\lvert m_i^{\text{subj}}-m_1^{\text{subj}}\rvert}{m_1^{\text{subj}}}.
\end{equation}

\subsubsection{Complementary Metrics}

Following previous minute-long generation works \citep{guo2025long,cai2025mixture}, we additionally include five complementary metrics from VBench \citep{huang2024vbench} that are essential for evaluating long video generation, including: (1) \textit{Imaging Quality}, which measures the technical fidelity of each video frame by quantifying distortions (e.g., over-exposure, noise, blur), thus reflecting the clarity and integrity of the generated imagery. (2) \textit{Motion Smoothness}, which assesses the fluidity and realism of movements in the video, ensuring that frame-to-frame transitions are continuous and physically plausible to achieve natural motion. (3) \textit{Aesthetic Quality}, which evaluates the visual appeal of the video frames, capturing artistic factors like composition, color harmony, photorealism, and overall beauty as perceived in each frame. 
(4) \textit{Background Consistency}, which measures the stability of the scene’s background across the video, determining whether the backdrop remains visually consistent throughout all frames. (5) \textit{Subject Consistency}, which evaluates whether a subject’s appearance remains consistent across every frame of the video, capturing the temporal coherence of that subject’s visual identity over the entire sequence.

\section{Limitations and Future Work} 
\label{sec:limitation}
While our framework performs well in single-shot long video generation, broader settings such as multi-shot composition remain to be explored, particularly regarding coherence across scene transitions. As future work, we aim to study these cases and consider extensions such as a larger InterVBench and 3D-aware modeling to further assess and broaden the method's applicability.

\subsection{Visualization Comparison}
\label{app:vis-compare}

The full prompts of the\textit{ Figure 2 in main paper} are as follows:

\begin{center}
\begin{tcolorbox}[promptbox, before skip=0pt, after skip=0pt]

"captions": [

"A serene white swan glides across a misty lake, its reflection shimmering in the calm water (00s - 03s).",

"The swan dips its head gracefully into the water,

creating gentle ripples around it (04s - 07s).",

"Lifting its head, the swan shakes off droplets, sending small splashes into the air (08s - 11s).",

"It spreads its wings slightly, flapping them to create a splash and adjust its position (12s - 15s).",

"The swan turns slightly, continuing to glide smoothly as mist hovers over the water (16s - 19s).",

"With elegant movements, the swan swims forward, its long neck curved gracefully (20s - 23s).",

"The swan pauses briefly, surveying its surroundings with a poised demeanor (24s - 27s).",

"It resumes swimming, its feathers catching the soft light filtering through the mist (28s - 31s).",

"Dipping its beak again, the swan appears to forage or drink from the tranquil waters (32s - 35s).",

"The swan lifts its head once more, shaking off water with a delicate motion (36s - 39s).",

"Turning its body, the swan reveals its full profile against the backdrop of foggy greenery (40s - 43s).",

"It continues its graceful journey, leaving a trail of ripples behind (44s - 47s).",

"The swan’s reflection mirrors its every move, enhancing the peaceful ambiance (48s - 51s).",

"As it drifts further away, the swan becomes part of the misty landscape (52s - 55s).",

"The swan slows down, almost still, embodying tranquility on the quiet lake (56s - 59s)."]
\end{tcolorbox}
\end{center}

As shown \textit{in Figure 2 in main paper}, all five baselines exhibit varying degrees of severe accumulation errors when generating minute-long videos. MAGI-I \citep{teng2025magi}, Self-Forcing \citep{huang2025self}, and PAVDM \citep{xie2025progressive} suffer from significant image quality degradation and color distortion after around 12 seconds, with the video gradually deteriorating and eventually collapsing. FramePack \citep{zhang2025packing}, on the other hand, avoids severe image distortion but produces poor dynamics and limited content diversity due to its symmetric progression design. SkyReel-V2 \citep{chen2025skyreels} is the closest baseline in comparison, yet it still experiences noticeable color drift after 12 seconds, which continues to accumulate until the final block. In contrast, our method outperforms all of these approaches, maintaining subject and background consistency, preserving image quality, and preventing color degradation.

\newpage

\begin{figure*}[h]
\centering
\includegraphics[width=\linewidth]{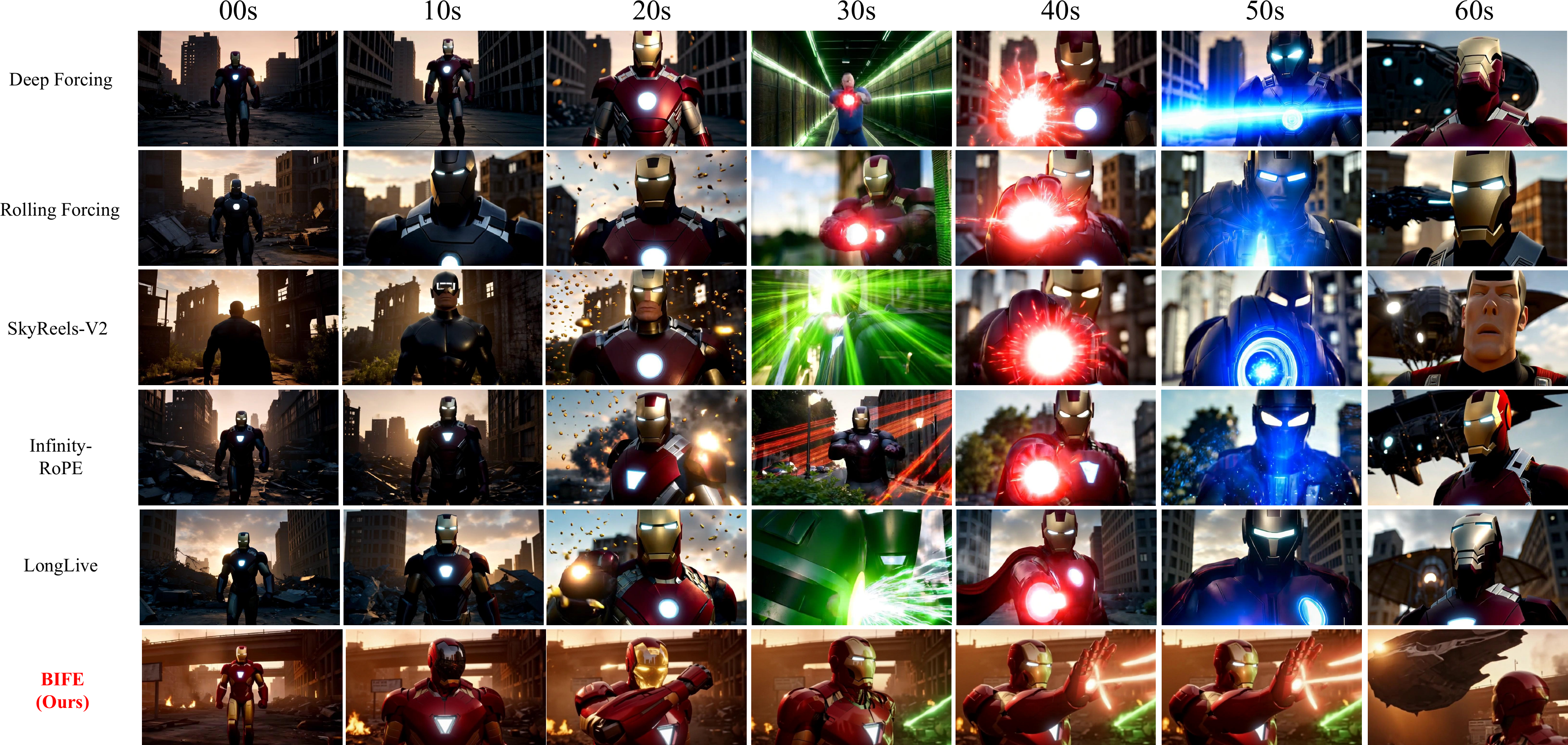}
\caption{An interactive long-video generation example.}
\label{fig:vis_1}
\end{figure*}

Moreover, Figures~\ref{fig:vis_1} presents qualitative comparisons on an interactive long-video generation example. 
Compared with existing approaches, BIFE maintains stronger subject consistency, smoother motion transitions, and more stable scene structure across the entire sequence. 

\begin{center}
\begin{tcolorbox}[promptbox, fontupper=\small, before skip=0pt, after skip=0pt]

"captions": [

"Iron Man walks through a war-torn, crumbling urban ruin at dusk, then stops and stands still. Wide shot to medium close-up (00s - 09s).",

"A hail of bullets tears in. Iron Man raises his arm, rounds and shells streak past. Wide shot to medium close-up (10s - 19s).",

"Emerald-green beams sweep down the street. Wide shot to medium close-up (20s - 29s).",

"Iron Man aims and fires tight, pulsed red repulsor blasts from his palm. Wide shot to medium close-up (30s - 39s).",

"The chest arc reactor unleashes a colossal blue beam. Wide shot to medium close-up (40s - 49s).",

"A dark alien craft enters from the left and cruises across. Iron Man raises his arm again while looking up at the alien craft. Wide shot to medium close-up (50s - 59s)."]

\end{tcolorbox}
\end{center}

\clearpage

\begin{figure*}[t]
    \centering
    \includegraphics[width=\linewidth]{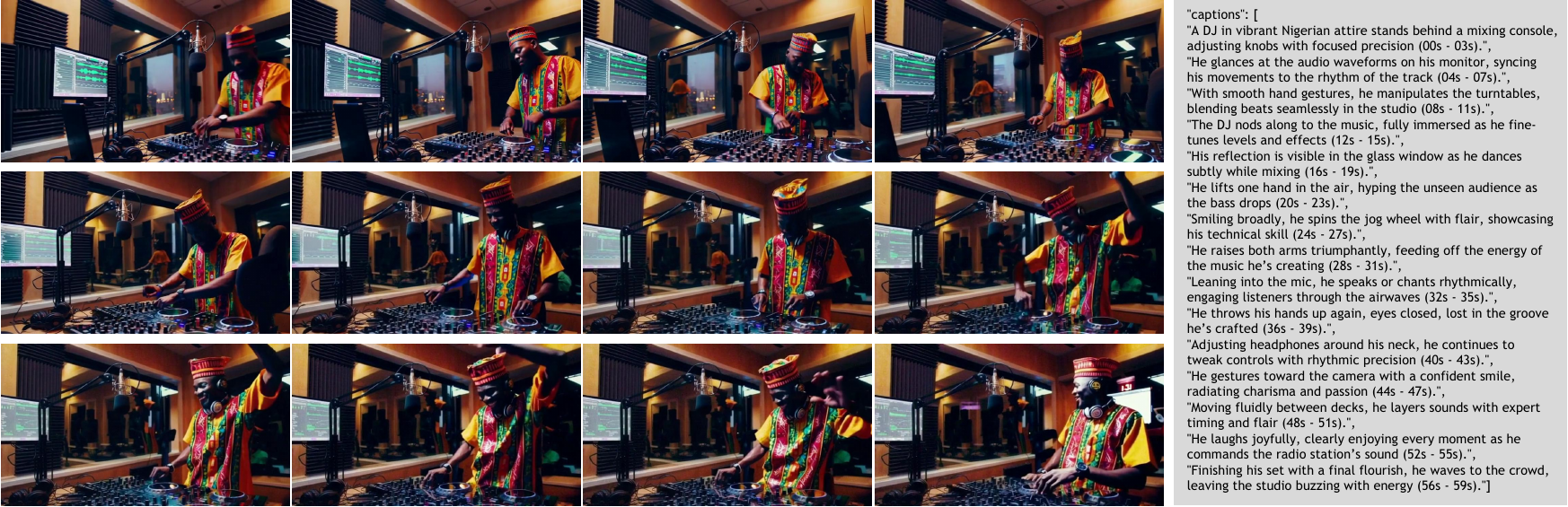}
    \caption{More visualization results \#1.}
    \label{fig:more-1}

    \vspace{3em}

    \includegraphics[width=\linewidth]{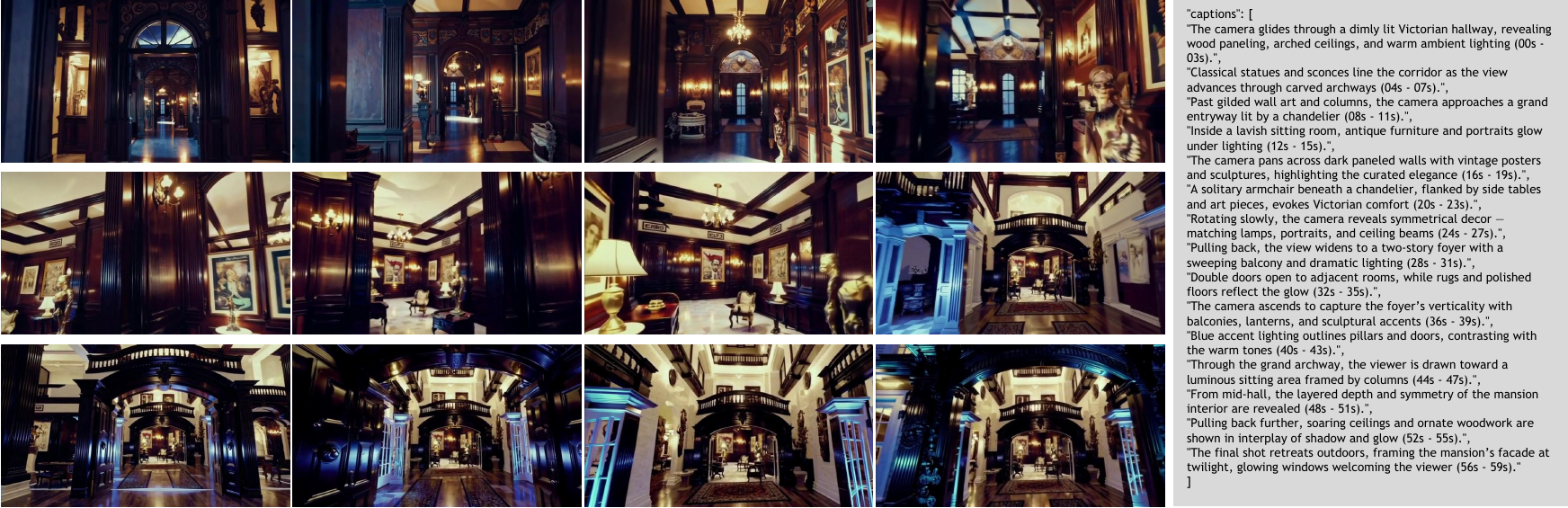}
    \caption{More visualization results \#2.}
    \label{fig:more-2}

    \vspace{3em}

    \includegraphics[width=\linewidth]{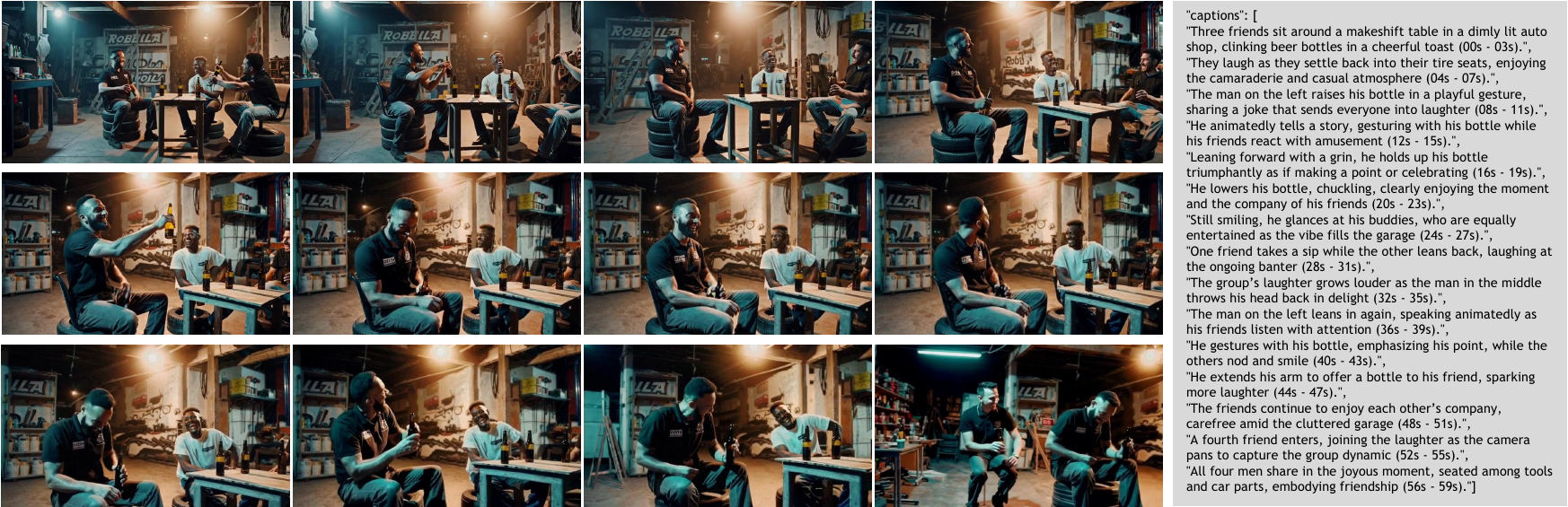}
    \caption{More visualization results \#3.}
    \label{fig:more-3}
\end{figure*}

\clearpage

\begin{figure*}[t]
    \centering
    \includegraphics[width=\linewidth]{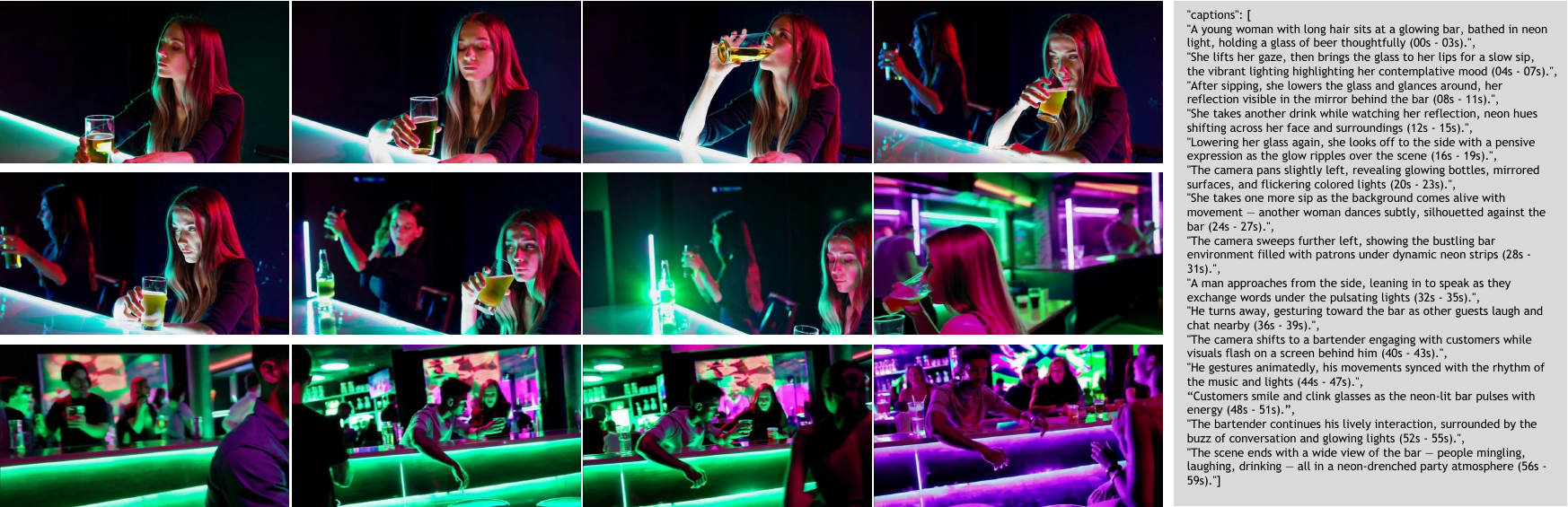}
    \caption{More visualization results \#4.}
    \label{fig:more-4}

    \vspace{3em}

    \includegraphics[width=\linewidth]{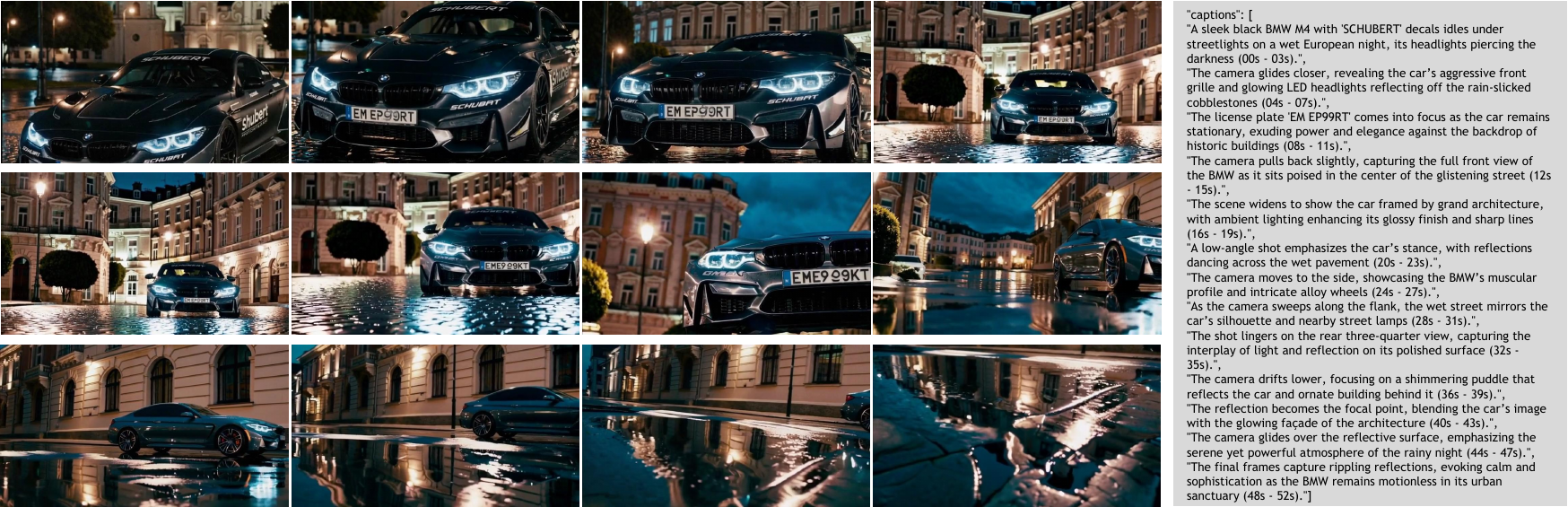}
    \caption{More visualization results \#5.}
    \label{fig:more-5}

    \vspace{3em}

    \includegraphics[width=\linewidth]{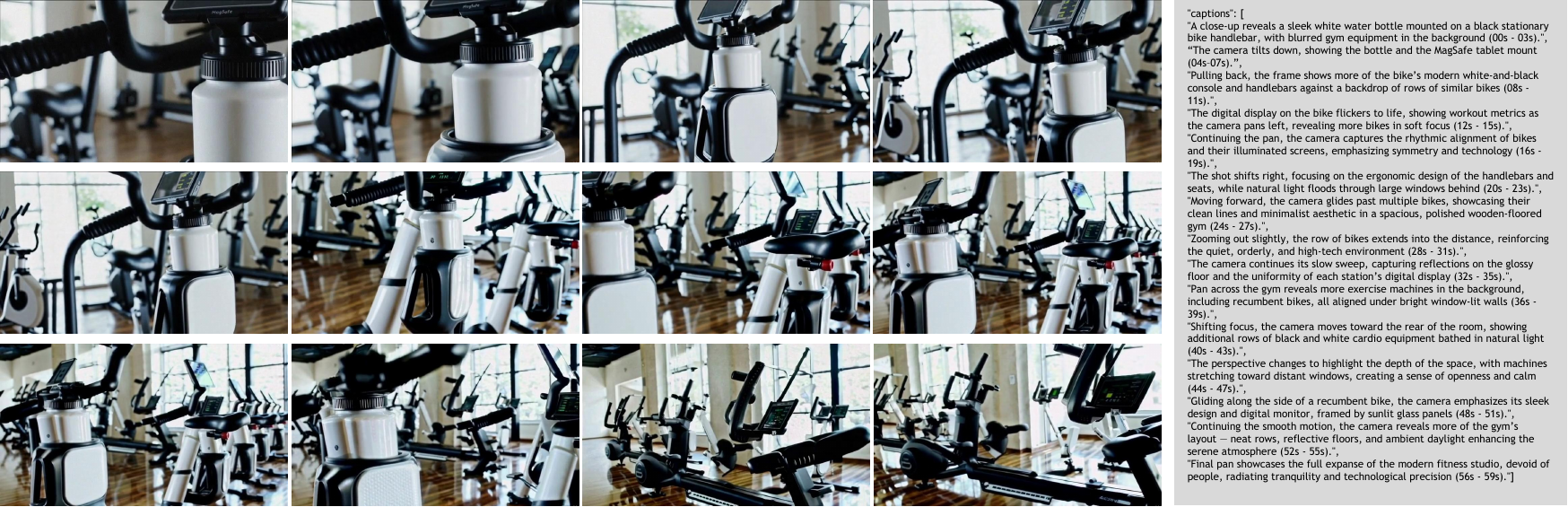}
    \caption{More visualization results \#6.}
    \label{fig:more-6}
\end{figure*}

\end{document}